\documentclass{article}

\usepackage{ijcai21}
\usepackage{comment}

\usepackage{times}
\usepackage{soul}
\usepackage{url}
\usepackage[hidelinks]{hyperref}
\usepackage[utf8]{inputenc}
\usepackage[small]{caption}
\usepackage{graphicx}
\usepackage{subcaption}
\usepackage{amssymb}
\usepackage{amsfonts}
\usepackage{amsmath}
\usepackage{amsthm}
\usepackage{booktabs}
\usepackage{algorithm}
\usepackage{algorithmic}
\usepackage{xcolor}
\usepackage{tabularx, booktabs} 
\usepackage{graphicx}
\usepackage{booktabs}
\usepackage{multirow}
\usepackage{colortbl}
\usepackage{xcolor}
\usepackage{amssymb}
\usepackage{pifont}
\usepackage{hyperref}
\usepackage{placeins}
\usepackage{pifont}

\usepackage[
  separate-uncertainty = true,
  multi-part-units = repeat
]{siunitx}

\definecolor{dgreen}{rgb}{0.0, 0.5, 0.0}
\newcommand{\cmark}{\color{dgreen} \ding{51}}%
\newcommand{\xmark}{\color{red} \ding{55}}%
\usepackage{lipsum}

\usepackage{changepage}

\newcommand{\etal}{\emph{et al}.}

\newcommand{\mcc}[1]{\multicolumn{#1}{c}}
\newcommand{\mcp}[1]{\multicolumn{#1}{c}} 
\newcommand{\B}[1]{\textcolor{black}{\textbf{#1}}}

\newcommand{\texta}[1]{$\texttt{#1}$}

\definecolor{Gray}{gray}{0.90}
\newcolumntype{b}{>{\columncolor{Gray}}c}
\newcolumntype{a}{>{\columncolor{Gray}}r}
\newcolumntype{d}{>{\columncolor{Gray}}c}

\title{
\color{black} 
Look Wide and Interpret Twice: 
Improving Performance on Interactive Instruction-following Tasks
}

\author{
Van-Quang Nguyen$^1$
\and
Masanori Suganuma$^{2,1}$\and
Takayuki Okatani$^{1,2}$
\affiliations
$^1$Graduate School of Information Sciences, Tohoku University \\
$^2$RIKEN Center for AIP\\

\emails
\texttt{\footnotesize \{quang,suganuma,okatani\}@vision.is.tohoku.ac.jp}
}

\begin{document}

\maketitle

\begin{abstract}
There is a growing interest in the community in making an embodied AI agent perform a complicated task while interacting with an environment following natural language directives. Recent studies have tackled the problem using ALFRED, a well-designed dataset for the task, but achieved only very low accuracy. This paper proposes a new method, which outperforms the previous methods by a large margin. It is based on a combination of several new ideas. One is a two-stage interpretation of the provided instructions. The method first selects and interprets an instruction without using visual information, yielding a tentative action sequence prediction. It then integrates the prediction with the visual information etc., yielding the final prediction of an action and an object. As the object's class to interact is identified in the first stage, it can accurately select the correct object from the input image. Moreover, our method considers multiple egocentric views of the environment and extracts essential information by applying
hierarchical attention conditioned on the current instruction. This contributes to the accurate prediction of actions for navigation. A preliminary version of the method won the ALFRED Challenge 2020. The current version achieves the unseen environment's success rate of {\bf4.45\%} with a single view, which is further improved to {\bf 8.37\%} with multiple views.
\end{abstract}

\section{Introduction}

There is a growing interest in the community in making an embodied AI agent perform a complicated task following natural language directives. Recent studies of 
vision-language navigation tasks (VLN) have made significant progress \cite{anderson2018vision,fried2018speaker,zhu2020vision}. However, these studies consider navigation in static environments, where the action space is simplified, and there is no interaction with objects in the environment.

To consider more complex tasks, a benchmark named ALFRED was developed recently \cite{shridhar2020alfred}. It requires an agent to accomplish a household task in interactive environments following given language directives. Compared with VLN, ALFRED is more challenging as the agent needs to (1) reason over a greater number of instructions and (2) predict actions from larger action space to perform a task in longer action horizons. The agent also needs to (3) localize the objects to manipulate by predicting the pixel-wise masks. Previous studies (e.g., \cite{shridhar2020alfred}) employ a Seq2Seq model, which performs well on the VLN tasks \cite{ma2019selfmonitoring}. However, it works poorly on ALFRED. Overall, existing methods only show limited performance; there is a huge gap with human performance. 

In this paper, we propose a new method that leads to significant performance improvements. It is based on several ideas. Firstly, we propose to choose a single instruction to process at each timestep from the given series of instructions. This approach contrasts with previous methods that encode them into a single long sequence of word features and use soft attention to specify which instruction to consider at each timestep implicitly \cite{shridhar2020alfred,legg2020eccv,singh2020eccv}. Our method chooses individual instructions explicitly by learning to predict when the agent completes an instruction. This makes it possible to utilize constraints on parsing instructions, leading to a more accurate alignment of instructions and action prediction. 

Secondly, we propose a two-stage approach to the interpretation of the selected instruction. In its first stage, the method interprets the  instruction without using visual inputs from the environment,  yielding a tentative prediction of an action-object sequence. In the second stage, the prediction is integrated with the visual inputs to predict the action to do and the object to manipulate. The tentative interpretation makes it clear to interact with what class of objects, contributing to an accurate selection of objects to interact with. 

Moreover, we acquire multiple agent egocentric views of a scene as visual inputs and integrate them using a hierarchical attention mechanism. This allows the agent to have a wider field of views, leading to more accurate navigation. To be specific, converting each view into an object-centric representation, we integrate those for the multiple views into a single feature vector using hierarchical attention conditioned on the current instruction. 

Besides, we propose a module for predicting precise pixel-wise masks of objects to interact with, referred to as the mask decoder. It employs the object-centric representation of the center view, i.e., multiple object masks detected by the object detector. The module selects one of these candidate masks to specify the object to interact with. In the selection,  self-attention is applied to the candidate masks to weight them; they are multiplied with the tentative prediction of the pairs of action and an object class and the detector's confidence scores for the candidate masks.

The experimental results show that the proposed method outperforms all the existing methods by a large margin and ranks first in the challenge leaderboard as of the time of submission. A preliminary version of the method won the ALFRED Challenge 2020 \footnote{The ALFRED Challenge 2020 \href{https://askforalfred.com/EVAL}{https://askforalfred.com/EVAL}}. The present version further improved the task success rate in unseen and seen environments to 8.37\% and 29.16\%, respectively, which are significantly higher than the previously published SOTA (0.39\% and 3.98\%, respectively) \cite{shridhar2020alfred}. 

\section{Related Work}

\subsection{Embodied Vision-Language Tasks}

Many studies have been recently conducted on the problem of making an embodied AI agent follow natural language directives and accomplish the specified tasks in a three-dimensional environment while properly interacting with it. Vision-language navigation (VLN) tasks have been the most extensively studied, which require an agent to follow navigation directions in an environment.

Several frameworks and datasets for simulating real-world environments have been developed to study the VLN tasks. The early ones lack photo-realism and/or natural language directions \cite{kempka2016vizdoom,ai2thor,wu2018building}. Recent studies consider perceptually-rich simulated environments and natural language navigation directions  \cite{anderson2018vision,chen2019touchdown,hermann2020learning}.
In particular, since the release of the Room-to-Room (R2R) dataset \cite{anderson2018vision} that is based on real imagery \cite{chang2017matterport3d}, VLN has attracted increasing attention, leading to the development of many methods \cite{fried2018speaker,wang2019reinforced,ma2019selfmonitoring,tan2019learning,majumdar2020improving}.

Several variants of VLN tasks have been proposed. A study \cite{Nguyen_2019_CVPR} allows the agent to communicate with an adviser using natural language to accomplish a given goal.          
In a study \cite{thomason2020vision}, the agent placed in an environment attempts to find a specified object by communicating with a human by natural language dialog. 
{
A recent study \cite{suhr-etal-2019-executing} proposes the interactive environments where users can collaborate with an agent by not only instructing it to complete tasks, but also acting alongside it.}
Another study \cite{krantz2020beyond} introduces a continuous environment based on the R2R dataset that enables an agent to take more fine-grained navigation actions.
A number of other embodied vision-language tasks have been proposed such as visual semantic planning \cite{zhu2017visual,gordon2019should} and embodied question answering \cite{embodiedqa,gordon2018iqa,wijmans2019embodied,puig2018virtualhome}.

\subsection{Existing Methods for ALFRED}
As mentioned earlier, ALFRED was developed to consider more complicated interactions with environments, which are missing in the above tasks, such as manipulating objects. Several methods for it have been proposed so far. A baseline method \cite{shridhar2020alfred} employs a Seq2Seq model with an attention mechanism and a progress monitor \cite{ma2019selfmonitoring}, which is prior art for the VLN tasks. In \cite{singh2020eccv}, a pre-trained Mask R-CNN is employed to generate object masks. It is proposed in \cite{legg2020eccv} to train the agent to follow instructions and reconstruct them. In \cite{corona2020modularity}, a modular architecture is proposed to exploit the compositionality of instructions. These methods have brought about only modest performance improvements over the baseline. A concurrent study \cite{singh2020moca} proposes a modular architecture design in which the prediction of actions and object masks are treated separately, as with ours. Although it achieves notable performance improvements, the study's ablation test indicates that the separation of the two is not the primary source of the improvements. {
Closely related to ALFRED, ALFWorld \cite{shridhar2021alfworld} has been recently proposed to combine TextWorld \cite{cote18textworld} and ALFRED for creating aligned environments, which enable transferring high-level policies learned in the text world to the embodied world.}

\section{Proposed Method}

The proposed model consists of three decoders (i.e., instruction, mask, and action decoders) with the modules extracting features from the inputs, i.e., the visual observations of the environment and the language directives. We first summarize ALFRED and then explain the components one by one.

\begin{figure*}[t!]
    \vspace{-1.0em}
    \centering
    \includegraphics[width=\linewidth]{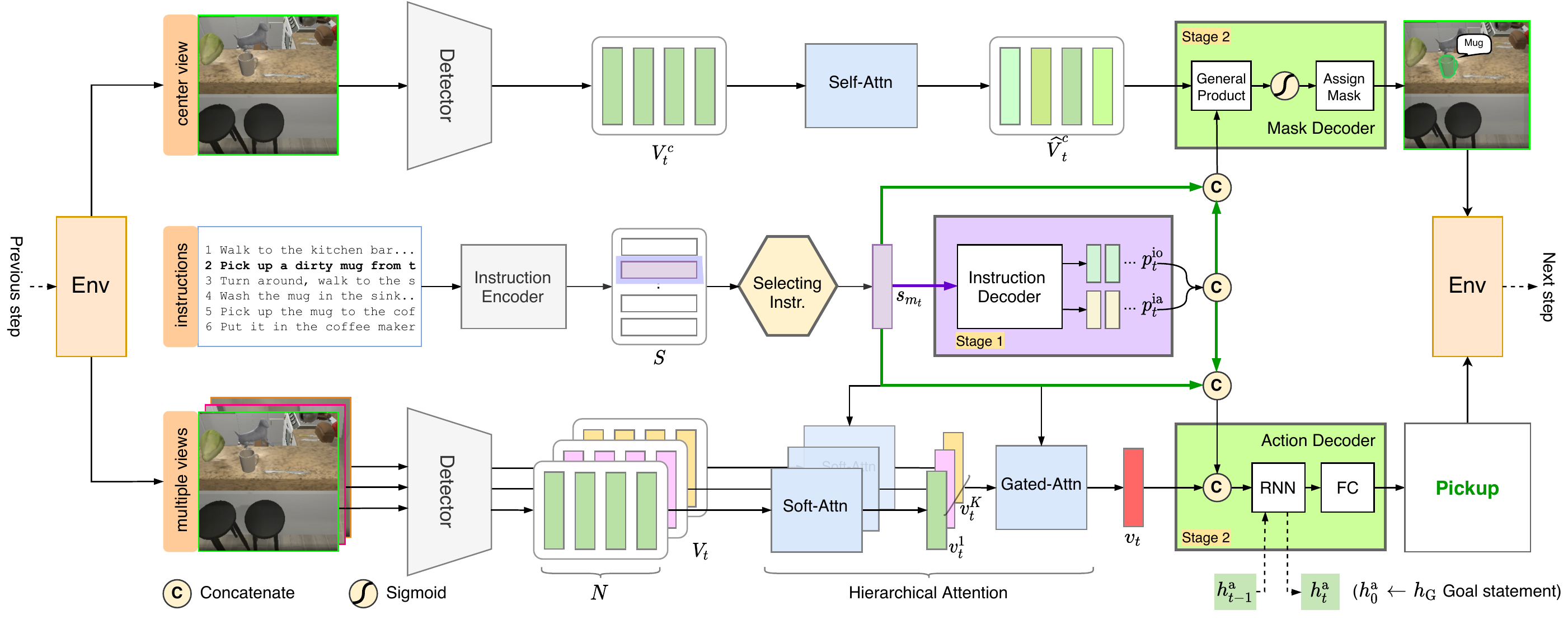}
    \caption{
        \textbf{Architecture overview of the proposed model.} It consists of the modules encoding the visual inputs and the language directives (Sec.~\ref{sec:feat_representation}), the instruction decoder with an instruction selector (Sec.~\ref{sec:inst_decoder}), the action decoder (Sec.~\ref{sec:action_decoder}), and the mask decoder (Sec.~\ref{sec:mask_decoder}).}         
    \vspace{-0.75em}
    \label{fig_overview}
\end{figure*}

\subsection{Summary of ALFRED}
ALFRED is built upon AI2Thor \cite{ai2thor}, a simulation environment for embodied AI. An agent performs seven types of tasks in 120 indoor scenes that require interaction with 84 classes of objects, including 26 receptacle object classes.
For each object class, there are multiple visual instances with different shapes, textures, and colors. 

The dataset contains 8,055 expert demonstration  episodes of task instances.
They are sequences of actions, whose average length is 50, and they are used as a ground truth action sequence at training time. For each of them, language directives annotated by AMT workers are provided, which consist of a goal statement $G$ and a set of step-by-step instructions, $S_{1},\ldots, S_{L}$. The alignment between each instruction and a segment of the action sequence is known. As multiple AMT workers annotate the same demonstrations, there are 25,743 language directives in total. 

We wish to predict the sequence of agent's actions, given $G$ and $S_1,\ldots,S_L$ of a task instance. 
There are two types of actions, navigation actions and manipulation actions. There are five navigation actions (e.g., \texta{MoveAhead} and \texta{RotateRight}) and seven manipulation actions (e.g., \texta{Pickup} and \texta{ToggleOn}). The manipulation actions accompany an object. The agent specifies it using a pixel-wise mask in the egocentric input image. Thus, the {\em outputs} are a sequence of actions with, if necessary, the object masks.

\subsection{Feature Representations} \label{sec:feat_representation}
\subsubsection{Object-centric Visual Representations}

Unlike previous studies \cite{shridhar2020alfred,singh2020eccv,legg2020eccv}, we employ the object-centric representations of a scene \cite{devin2018deep}, which are extracted from a pretrained object detector (i.e., Mask R-CNN \cite{he2017mask}). It provides richer spatial information about the scene at a more fine-grained level and thus allows the agent to localize the target objects better. Moreover, we make the agent look wider by capturing the images of its surroundings, aiming to enhance its navigation ability.

Specifically, at timestep $t$, the agent obtains visual observations from $K$ egocentric views. For each view $k$, we encode the visual observation
by a bag of $N$ object features,
which are extracted the object detector. Every detected object 
is associated with a visual feature, a mask, and its confidence score. We project the visual feature
into $\mathbb{R}^{d}$ with a linear layer, followed by a ReLU activation and dropout regularization \cite{srivastava2014dropout} to obtain a single vector; thus, we get 
a set of $N$ object features for view $k$, $V^{k}_{t} = (v^{k}_{t,1}, \ldots, v^{k}_{t,N})$. 
We obtain $V^{1}_{t}, \ldots, V^{K}_{t}$ 
for all the views.

\subsubsection{Language Representations}
We encode the language directives as follows. We use an embedding layer initialized with pretrained GloVe \cite{pennington2014glove} vectors to embed each word of the $L$ step-by-step instructions and the goal statement. For each instruction $i(=1,\ldots,L)$, the embedded feature sequence is inputted to a two-layer LSTM \cite{hochreiter1997long}, and its last hidden state is used as the feature $s_i\in\mathbb{R}^d$ of the instruction. We use the same LSTM for all the instructions with dropout regularization. We encode the goal statement $G$ in the same manner using an LSTM with the same architecture different weights, obtaining $h_\text{G}\in\mathbb{R}^d$.

\subsection{Instruction Decoder} \label{sec:inst_decoder} 

\subsubsection{Selecting Instructions}
Previous studies \cite{shridhar2020alfred,singh2020eccv,legg2020eccv} employ a Seq2Seq model in which all the language directives are represented as a {\em single sequence} of word features, and soft attention is generated over it to specify the portion to deal with at each timestep. 
We think this method could fail to correctly segment instructions with time, even with the employment of progress monitoring \cite{ma2019selfmonitoring}.
This method does not use a few constraints on parsing the step-by-step instructions that they should be processed in the given order and when dealing with one of them, the other instructions, especially the future ones, will be of little importance.

We propose a simple method that can take the above constraints into account, which explicitly represents which instruction to consider at the current timestep $t$. The method introduces an integer variable $m_t(\in[1,L])$ storing the index of the instruction to deal with at $t$. 

To update $m_t$ properly, we introduce a virtual action representing the {\em completion of a single instruction}, which we treat equally to the original twelve actions defined in ALFRED. Defining a new token \texttt{COMPLETE} to represent this virtual action, we augment each instruction's action sequence provided in the expert demonstrations 
always end with \texttt{COMPLETE}.
At training time, we train the action
decoder to predict the augmented sequences. At test time, the same decoder predicts an action at each timestep; if it predicts \texttt{COMPLETE}, this means completing the current instruction. 
The instruction index $m_t$ is updated as follows:%
\begin{equation}
    m_t= 
    \begin{cases}
        m_{t-1} + 1,& \text{if } \mathrm{argmax}(p^\text{a}_{t-1}) = \texttt{COMPLETE}\\
        m_{t-1},              & \text{otherwise},
    \end{cases}
\end{equation}
where $p^\text{a}_{t-1}$ is the predicted probability distribution over all the actions at time $t-1$, which will be explained in Sec.~\ref{sec:action_decoder}. 
The encoded feature $s_{m_t}$ of the selected instruction is used in all the subsequent components, as shown in Fig.~\ref{fig_overview}.

\begin{figure}[!ht]
    \centering
    \includegraphics[width=1.0\linewidth]{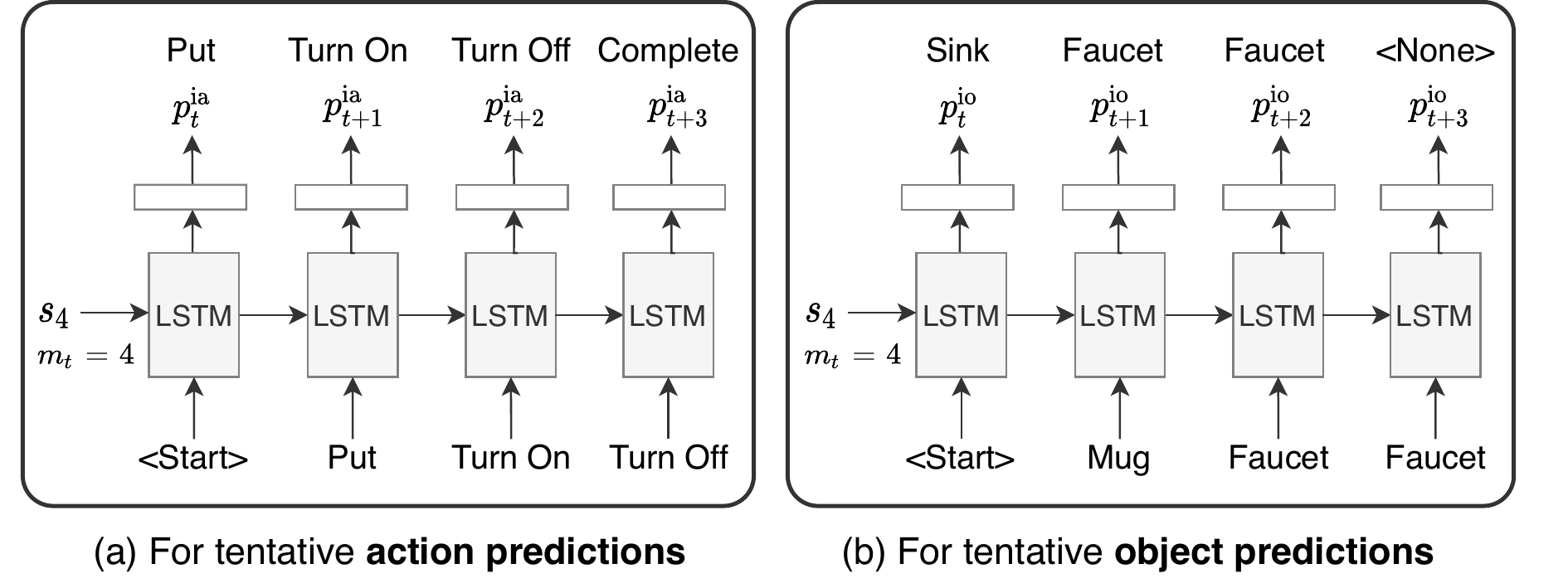}
    \caption{
        {
        An example illustrates how we reinitialize the hidden states of the two LSTMs in the instruction encoder by $s_{m_t}$ when $m_t = m_{t-1} + 1$ ($m_t = 4)$.}}    
    \label{fig:inst_decoder}
    \vspace{-0.2cm}
\end{figure}
\subsubsection{Decoder Design} 
As explained earlier, our method employs a two-stage approach for interpreting the instructions. The instruction decoder (see Fig.~\ref{fig_overview}) runs the first stage, where it interprets the instruction encoded as $s_{m_t}$ {\em without any visual input}. To be specific, it transforms $s_{m_t}$ into the sequence of action-object pairs without additional input. In this stage, objects mean the {\em classes} of objects. 

As it is not based on visual inputs, the predicted action-object sequence has to be tentative. The downstream components in the model (i.e., the mask decoder and the action decoder) interpret $s_{m_t}$ again, yielding the final prediction of an action-object sequence, which are grounded on the visual inputs. Our intention of this two-stage approach is to increase prediction accuracy; we expect that using a prior prediction of (action, object class) pairs helps more accurate grounding.

In fact, many instructions in the dataset, particularly those about interactions with objects, are sufficiently specific so that they are uniquely translated into (action, object class) sequences with a perfect accuracy, even without visual inputs. For instance,
``
Wash the mug in the sink'' can be  translated into 
(\texta{Put}, \texta{Sink}), (\texta{TurnOn}, \texta{Faucet}), 
(\texta{TurnOff}, \texta{Faucet}), (\texta{PickUp}, \texta{Mug}).
However, this is not the case with navigation instructions. 
For instance, 
``{Go straight to the sink}'' may be translated into a variable number of repetition of \texta{MoveAhead}; it is also hard to translate ``{Walk into the drawers}'' when it requires to navigate to the left/right. Therefore, we separately deal with the manipulation actions and the navigation actions. In what follows, we first explain the common part and then  the different parts.

Given the encoded feature $s_{m_t}$ of the selected instruction, the instruction decoder predicts the action and the object class to choose at $t$. To be precise, it outputs the probability distributions $p^\text{ia}_t(\in\mathbb{R}^{N_\text{a}})$ and $p^\text{io}_t(\in\mathbb{R}^{N_\text{o}})$ over all the actions and the object classes, respectively; $N_\text{a}$ and $N_\text{o}$ are the numbers of the actions and the object classes. 

These probabilities $p^\text{ia}_t$ and $p^\text{io}_t$ are predicted separately by two LSTMs in an autoregressive fashion.
The two LSTMs are initialized whenever a new instruction is selected; to be precise, we reset their internal states as $h^\text{ia}_{t-1}=h^\text{io}_{t-1}=s_{m_t}$ for $t$ when we increment $m_t$ as $m_t=m_{t-1}+1$ ({
see the example in Fig. \ref{fig:inst_decoder}}). Then, $p^\text{ia}_t$ and $p^\text{io}_t$  are predicted as follows:
\begin{subequations}
\begin{align}
        p^\text{ia}_{t} &= \mathrm{softmax}(W_\text{ia}\text{LSTM}(E_{\text{a}}(p^\text{ia}_{t-1}), h^\text{ia}_{t-1}) + b_\text{ia}),\\ 
        p^\text{io}_{t} &= \mathrm{softmax}(W_\text{io}\text{LSTM}(E_{\text{o}}(p^\text{io}_{t-1}), h^\text{io}_{t-1}) + b_\text{io}), 
\end{align}
\end{subequations}
where $W_\text{ia} \in \mathbb{R}^{N_\text{a} \times d}$, 
$b_\text{ia}\in \mathbb{R}^{N_\text{a}}$, 
$W_\text{io} \in \mathbb{R}^{N_\text{o} \times d}$, 
and $b_\text{io}\in \mathbb{R}^{N_\text{o}}$ are learnable parameters; 
$E_\text{a}$ maps the most likely action into the respective vectors according to the last predictions $p^\text{ia}_{t-1}$ using a dictionary with $N_\text{a}\times d$ learnable parameters; $E_\text{o}$ does the same for the object classes. The predicted $p^\text{ia}_{t}$ and $p^\text{io}_{t}$ are transferred to the input of these LSTMs at the next timestep and also inputted to the downstream components, the mask decoder and the action decoder. 

Now, as they do not need visual inputs, we can train the two LSTMs in a supervised fashion using the pairs of instructions and the corresponding ground truth action-object sequences. We denote this supervised loss, i.e., the sum of the losses for the two LSTMs, by $\mathcal{L}_{\text{aux}}$. Although it is independent of the environment and we can train the LSTMs offline, we simultaneously train them along with other components in the model by adding $\mathcal{L}_{\text{aux}}$ to the overall loss. We think this contributes to better learning of instruction representation 
$s_{m_t}$, which is also used by the mask decoder and the action decoder.

As mentioned above, we treat the navigation actions differently from the manipulation actions. There are three differences. First, we simplify the ground truth action sequence for the navigation actions if necessary. For instance, suppose an instruction ``{Turn left, go ahead to the counter and turn right}'' with a ground truth action sequence ``\texta{RotateLeft}, \texta{MoveAhead}, \texta{MoveAhead}, \texta{MoveAhead}, \texta{MoveAhead}, \texta{RotateRight}''. The repetition of \texta{MoveAhead} reflects the environment and cannot be predicted without visual inputs. Thus, by eliminating the repeated actions, we convert the sequence into the minimum-length one, ``\texta{RotateLeft}, \texta{MoveAhead}, \texta{RotateRight}'', and regard it as the ground truth sequence, training the instruction decoder. Second, as there is no 
accompanied object for the  navigation actions, we use the object-class sequence ``\texta{None, None, None}'' as the ground truth. Third, in the case of navigation actions, we do not transfer the outputs $p^\text{ia}_{t}$ and $p^\text{io}_{t}$ to the mask decoder and the action decoder and instead feed constant (but learnable) vectors $p^\text{ia}_{\text{nav}} \in \mathbb{R}^{N_\text{a}}$ and $p^\text{io}_{\text{nav}}  \in \mathbb{R}^{N_\text{o}}$ to them. 
As the instruction decoder learns to predict the minimum-length action sequences as above, providing such predictions will be harmful for the action decoder. We avoid this by feeding $p^\text{ia}_{\text{nav}}$ and $p^\text{io}_{\text{nav}}$.

\subsection{Action Decoder} \label{sec:action_decoder}

The action decoder receives four inputs and predicts the action at $t$. The inputs are as follows: the encoded instruction $s_{m_t}$, the output $p_t^\text{ia}$ and $p_t^\text{io}$ of the instruction decoder\footnote{These are replaced with $p_{\text{nav}}^\text{ia}$ and $p_{\text{nav}}^\text{ia}$ if $\mathrm{argmax}(p^\text{ia}_t)$ is not a manipulation action, as mention above.}
and aggregated feature $v_t$ of visual inputs, which will be described below.

\subsubsection{Hierarchical Attention over Visual Features} 

As explained in Sec.~\ref{sec:feat_representation}, we use the multi-view object-centric representation of visual inputs. To be specific, we aggregate $N\times K$ outputs of Mask R-CNN from $K$ ego-centric images, obtaining a single vector $v_t$. The Mask R-CNN outputs for view $k(=1,\ldots,K)$ are the visual features  $(v^k_{t,1},\ldots,v^k_{t,N})$ and 
the confidence scores 
$(\rho^k_{t,1},\ldots,\rho^k_{t,N})$ of $N$ detected objects.

To do this feature aggregation, we employ a hierarchical approach, where we first search for the objects relevant to the current instruction in each view and then merge the features over the views to a single feature vector. In the first step, we compute and apply soft-attentions over $N$ objects for each view. To be specific, we compute attention weights
${\alpha}^{k}_{\text{s}} \in \mathbb{R}^N$ across
$v^{k}_{t,1},\ldots,v^{k}_{t,N}$ 
guided by $s_{m_t}$ as
\begin{equation}
    \alpha^{k}_{\text{s},n} = \mathrm{softmax}(({v}^{k}_{t,n})^\top W^k_\text{s} s_{m_t}), \\
\end{equation}
where $W^k_\text{s} \in \mathbb{R}^{d \times d}$ is a learnable matrix, for $k=1,\ldots,K$. 
We then apply the weights to the $N$ visual features multiplied with their confidence scores for this view, yielding a single $d$-dimensional vector as 
\begin{equation}
    {v}^{k}_{t} = \sum_{n=1}^{N} {\alpha}^{k}_{\text{s},n} 
    {v}^{k}_{t,n} \rho^k_{t,n},
\end{equation}
where $\rho^k_{t,n}$ is the confidence score associated with
$v^k_{t,n}$.

In the second step, we merge the above features $v^1_t,\ldots,v^K_t$ using 
\emph{gated-attention}. We compute the weight $\alpha_{g}^k (\in \mathbb{R})$ 
of view $k(=1,\ldots,K)$ guided by $s_{m_t}$ as
\begin{equation}
\label{eq:gate-attn}
    \alpha^{k}_{\text{g}} = \mathrm{sigmoid}(({v}^{k}_{t})^\top W_\text{g} s_{m_t}),
\end{equation}
where $W_\text{g}\in \mathbb{R}^{d \times d}$ is a learnable matrix.
Finally, we apply the weights to $\{v^k_t\}_{k=1,\ldots,K}$ to have the visual feature $v_t \in \mathbb{R}^d$ as
\begin{equation}
    v_t = \sum_{k=1}^{K} {\alpha}^{k}_{\text{g}} {v}^{k}_{t}.
\end{equation}
As shown in the ablation test in the appendix,
the performance drops significantly when replacing the above gated-attention by soft-attention, indicating the necessity for merging observations of different views, not selecting one of them. 

\subsubsection{Decoder Design}

The decoder predicts the action at $t$ from $v_t$, $s_{m_t}$, $p_t^\text{ia}$ and $p_t^\text{io}$. We employ an LSTM, which outputs the hidden state $h^\text{a}_{t}\in\mathbb{R}^d$ at $t$ from the previous state $h^\text{a}_{t-1}$ along with the above four inputs as
\begin{equation}\label{eq:action_pred}
    h^\text{a}_{t} = \mathrm{LSTM}([v_t; s_{m_t}; p^\text{ia}_{t}; p^\text{io}_{t}], h^\text{a}_{t-1}),
\end{equation}
where $[;]$ denotes concatenation operation. We initialize the LSTM by setting the initial hidden state $h^\text{a}_0$ to  $h_\text{G}$, the encoded feature of the goal statement; see Sec.~\ref{sec:feat_representation}. The updated state $h^\text{a}_{t}$ is fed into a fully-connected layer to yield the probabilities over the $N_\text{a} +1$ actions including \texttt{COMPLETE} as follows:
\begin{equation}
    p^\text{a}_t = \mathrm{softmax}(W_\text{a} h^\text{a}_{t}  + b_\text{a}),
\end{equation}
where $W_\text{a} \in \mathbb{R}^{(N_\text{a} + 1)\times d}$
and 
$b_\text{a} \in \mathbb{R}^{N_\text{a} + 1}$.
We choose the action with the maximum probability for the predicted action. In the training of the model, we use cross entropy loss $\mathcal{L}_\text{action}$ computed between $p^\text{a}_t$ and the one-hot representation of the true action.

\subsection{Mask Decoder} \label{sec:mask_decoder}

To predict the mask specifying an object to interact with, we utilize the object-centric representations 
$V^c_t=(v^c_{t,1},\ldots,v^c_{t,N})$
of the visual inputs of the central view ($k=c$).
Namely, we have only to select one of the $N$ detected objects. This enables more accurate specification of an object mask than predicting a class-agnostic binary mask as in the prior work \cite{shridhar2020alfred}.

To do this, we first apply simple self-attention to the visual features $V^c_t$, aiming at capturing the relation between objects in the central view. We employ the attention mechanism inside the light-weight Transformer with a single head proposed in \cite{nguyenefficient} for this purpose, obtaining $\bar{\cal A}_{V^c_t}(V^c_t) \in \mathbb{R}^{N\times d}$. We then apply linear transformation to $\bar{\cal A}_{V^c_t}(V^c_t)$ 
using a single fully-connected layer having weight $W\in \mathbb{R}^{N\times d}$ and bias $b\in \mathbb{R}^d$, 
with a residual connection as 
\begin{equation}
    \hat{V^c_t} = \mathrm{ReLU}(W \bar{\cal A}_{V^c_t}
    (V^c_t) +\mathbf{1}_{K} \cdot b^\top) + V^c_t,
\end{equation}
where $\mathbf{1}_K$ is  $K$-vector with all ones.

We then compute the probability $p^\text{m}_{t,n}$ of selecting $n$-th object from the $N$ candidates using the above self-attended object features along with other inputs $s_{m_t}$, $p^\text{ia}_{t}$, and $p^\text{io}_{t}$. We concatenate the latter three inputs into a vector $g^\text{m}_t = [s_{m_t}; p^\text{ia}_{t}; p^\text{io}_{t}]$ and then compute the probability as 
\begin{equation}\label{eq:mask_pred}
    p^\text{m}_{t,n} = \mathrm{sigmoid}((g^\text{m}_t)^\top  W_\text{m} \hat{v}^{c}_{t,n}),
\end{equation}
where $W_\text{m}\in\mathbb{R}^{d + N_\text{a} + N_\text{o} \times d}$ is a learnable matrix. We select the object mask with the highest probability (i.e., $\mathrm{argmax}_{n=1,\ldots,N}(p^\text{m}_{t,n})$) at inference time.
At training time, we first match the ground truth object mask with the object mask having the highest IoU. Then, we calculate the BCE loss $\mathcal{L}_\text{mask}$ between the two. 

\section{Experiments}

\subsection{Experimental Configuration}

\paragraph{Dataset.}

We follow the standard procedure of ALFRED; 25,743 language directives over 8,055 expert demonstration episodes are split into the training, validation, and test sets. The latter two are further divided into two splits, called {\em seen} and {\em unseen}, depending on whether the scenes are included in the training set. 

\paragraph{Evaluation metrics.}
Following ~\cite{shridhar2020alfred}, we report the standard metrics, i.e.,  the scores of Task Success Rate, denoted by \textbf{Task} and Goal Condition Success Rate, denoted by \textbf{Goal-Cond}. The \textbf{Goal-Cond} score is the ratio of goal conditions being completed at the end of an episode. The \textbf{Task} score is defined to be one if all the goal conditions are completed, and otherwise 0. Besides, each metric is accompanied by a path-length-weighted (PLW) score \cite{Anderson2018OnEO}, which measures the agent's efficiency by penalizing scores with the length of the action sequence.

\paragraph{Implementation details.}

\begin{table*}[t!]
    \centering
    \resizebox{1.00\textwidth}{!}{
        \begin{tabular}{@{}laarrcaarr@{}}
            \toprule
            \multicolumn{1}{l}{\multirow{3}{*}{Model}}
                             & \mcp{4}{\textbf{Validation}} & & \mcc{4}{\textbf{Test}} \\
                             
                             & \mcc{2}{\textit{Seen}}   & \mcc{2}{\textit{Unseen}}
                             & 
                             & \mcc{2}{\textit{Seen}}   & \mcc{2}{\textit{Unseen}}  \\
                             
                             & \multicolumn{1}{b}{Task} & \multicolumn{1}{b}{Goal-Cond} 
                             & \multicolumn{1}{c}{Task} & \multicolumn{1}{c}{Goal-Cond} 
                             & 
                             & \multicolumn{1}{b}{Task} & \multicolumn{1}{b}{Goal-Cond} 
                             & \multicolumn{1}{c}{Task} & \multicolumn{1}{c}{Goal-Cond} \\
            \cmidrule{1-5} \cmidrule{7-10}
            \multicolumn{8}{l}{\bf Single view} \\
            \multicolumn{1}{l}{\cite{shridhar2020alfred}}   & $3.70$ ($2.10$)    & $10.00$  ($7.00$)    & $0.00$ ($0.00$)   & $6.90$ ($5.10$)  & & $3.98$ ($2.02$)   & $9.42$ ($6.27$)   & $0.39$ ($0.80$) & $7.03$ ($4.26$) \\[1pt]            
            \multicolumn{1}{l}{\cite{legg2020eccv}}          &\multicolumn{1}{b}{-} & \multicolumn{1}{b}{-} & \multicolumn{1}{c}{-} & \multicolumn{1}{c}{-}     & & ${3.85}$ (${1.50}$)   & ${8.87}$ (${5.52}$)   & ${0.85}$ (${0.36}$) & ${7.68}$ (${4.31}$)        \\[1pt]
            \multicolumn{1}{l}{\cite{singh2020eccv}}        & ${4.50}$ (${2.20}$)    & ${12.20}$  (${8.10}$)    & ${0.70}$ (${0.30}$)   & ${9.50}$ (${6.10}$)        & & ${5.41}$ (${2.51}$)   & ${12.32}$ (${8.27}$)   & ${1.50}$ (${0.7}$) & ${8.08}$ (${5.20}$)        \\[1pt]
            \multicolumn{1}{l}{\cite{singh2020moca}}        & ${19.15}$ (${13.60}$)    & ${28.50}$  (${22.30}$)    & ${3.78}$ (${2.00}$)   & ${13.40}$ (${8.30}$)    & & ${22.05}$ (${15.10}$)   & ${28.29}$ (${22.05}$)   & ${5.30}$ (${2.72}$) & ${14.28}$ (${9.99}$) \\[1pt]
            \multicolumn{1}{l}{Ours (1 visual view)}& ${18.90}$ (${13.90}$)    & ${26.80}$  (${21.90}$)     & ${3.90}$ (${2.50}$)    & ${15.30}$ (${10.90}$)      & & $15.20$ ($11.79$) & $23.95$ ($20.27$)   & $4.45$ ($2.37$) & $14.71$ ($10.88$) \\[1pt]

            \cmidrule{1-5} \cmidrule{7-10}
            \multicolumn{8}{l}{\bf Multiple views} \\
            \multicolumn{1}{l}{Ours (5 visual views)}              & {$\B{33.70}$ ($\B{28.40}$)}    & $\B{43.10}$  ($\B{38.00}$)    & $\B{9.70}$ ($\B{7.30}$)    & $\B{23.10}$ ($\B{18.10}$)      & & $\B{29.16}$ ($\B{24.67}$) & $\B{38.82}$ ($\B{34.85}$)   & $\B{8.37}$ ($\B{5.06}$) & $\B{19.13}$ ($\B{14.81}$) \\[1pt]
            \multicolumn{1}{l}{Ours (5 visual views)$^\diamond$}   & ${14.30}$ (${10.80}$)    & ${22.40}$  (${19.60}$)     & ${4.60}$ (${2.80}$)    & ${11.40}$ (${8.70}$)         & & $12.39$ ($8.20$) & $20.68$ ($18.79$)   & $4.45$ ($2.24$) & $12.34$ ($9.44$) \\[1pt]

            \cmidrule{1-5}\cmidrule{7-10}
            \multicolumn{1}{l}{Human}     & \multicolumn{1}{b}{-} & \multicolumn{1}{b}{-} & \multicolumn{1}{c}{-} & \multicolumn{1}{c}{-} & & \multicolumn{1}{b}{-} & \multicolumn{1}{b}{-} & $91.00$ ($85.80$) & $94.50$ ($87.60$) \\
            \bottomrule
        \end{tabular}
    }
    \caption{\textbf{Task and Goal-Condition Success Rate.}
        For each metric, the corresponding path weighted metrics are given in (parentheses).
        The highest values per fold and metric are shown in \B{bold}. Our winning entry in the ALFRED Challenge 2020 is denoted with $^\diamond$ .
    }
    \label{tab:results}
\end{table*}

We use $K=5$ views:
the center view, \emph{up} and \emph{down} views with the elevation degrees of $\pm 15^{\circ}$, and \emph{left} and \emph{right} views with the angles of $\pm 90^{\circ}$.
We employ a Mask R-CNN model with ResNet-50 backbone that receives a $300 \times 300$ 
image and outputs $N = 32$ 
object candidates. 
We train it before training the proposed model with 800K frames and corresponding instance segmentation masks collected by replaying the expert demonstrations of the training set. We set the feature dimensionality $d=512$.
We train the model using imitation learning on the expert demonstrations by minimizing the following loss:
\begin{equation}
    \mathcal{L} = \mathcal{L}_\text{mask} + \mathcal{L}_\text{action} + 
    \mathcal{L}_\text{aux}.
\end{equation}
We use the Adam optimizer with an initial learning rate of $10^{-3}$, which is halved at epoch 5, 8, and 10, and a batch size of 32 for 15 epochs in total. We use a dropout with the dropout probability 0.2 for the both visual features and LSTM decoder hidden states.

\subsection{Experimental Results} \label{sec:result}

\begin{table}[t!]
    \centering
    \resizebox{0.48\textwidth}{!}{
        \begin{tabular}{@{}ldcdcdcdc@{}}
        \toprule
        \multirow{2}{*}{Sub-goal}
                         & \multicolumn{2}{c}{\small\cite{shridhar2020alfred}}
                         & \multicolumn{2}{c}{\small\cite{singh2020moca}} 
                         & \multicolumn{2}{c}{\textbf{Ours}}
                         \\
                          \cmidrule(lr){2-3} \cmidrule(lr){4-5}  \cmidrule(lr){6-7}
                         & \multicolumn{1}{a}{Seen} & \multicolumn{1}{c}{Unseen} 
                         & \multicolumn{1}{a}{Seen} & \multicolumn{1}{c}{Unseen}
                         & \multicolumn{1}{a}{Seen} & \multicolumn{1}{c}{Unseen}
                         \\
                         
        \hline
        {Goto}   & $51$ & $22$ & ${54}$ & $ {32}$ & $\B{59}$ & $ \B{39}$ \\
        \hline
        {Pickup} & $32$ & $21$ & ${53}$ & $ {44}$ & $\B{84}$ & $ \B{79}$ \\
        {Put}    & ${81}$ & ${46}$ & $62$ & $39$  & $\B{82}$ & $ \B{66}$ \\
        {Slice}  & $25$ & $12$ & ${51}$ & ${55}$  & $\B{89}$ & $ \B{85}$ \\
        \hline
        {Cool}   & ${88}$ & ${92}$ & $87$ & $38$  & $\B{92}$ & $ \B{94}$ \\
        {Heat}   & ${85}$ & ${89}$ & $84$ & $86$  & $\B{99}$ & $ \B{95}$ \\
        {Clean}  & ${81}$ & $57$ & $79$ & $\B{71}$  & $\B{94}$ & $ {68}$ \\
        {Toggle} & $\B{100}$ & ${32}$ & $93$ & $11$ & ${99}$ & $ \B{66}$ \\
        \hline
        {Average}    & $68$ & $46$ & ${70}$ & ${47}$ &\B{87} & \B{74} \\
        \bottomrule
        \end{tabular}
    }

    \caption{
        \textbf{Sub-goal success rate.} All values are in percentage. The agent is evaluated on the Validation set. Highest values per fold are indicated in \B{bold}.
    }
    \label{tab:res_subgoal}
\end{table}

Table~\ref{tab:results} shows the results. It is seen that our method shows significant improvement over the previous methods~\cite{shridhar2020alfred,legg2020eccv,singh2020eccv,singh2020moca} on all metrics. Our method also achieves better PLW (path length weighted) scores in all the metrics (indicated in the parentheses), showing its efficiency. Notably, our method attains \textbf{8.37\%} success rate on the unseen test split, improving approximately 20 times compared with the published result in \cite{shridhar2020alfred}. The higher success rate in the unseen scenes indicates its ability to generalize in novel environments. Detailed results for each of the seven task types are shown in the appendix.

The preliminary version of our method won an international competition, whose performance is lower than the present version. It differs in that $(p^\text{ia}_{t}, p^\text{io}_{t})$ are not forwarded to the mask decoder and the action decoder and the number of Mask R-CNN's outputs is set to $N=20$. It is noted that even with a single view (i.e., $K=1$), our model still outperforms  ~\cite{shridhar2020alfred,legg2020eccv,singh2020eccv} in all the metrics.

\paragraph{Sub-goal success rate.}

Following~\cite{shridhar2020alfred}, we evaluate the performance on individual sub-goals.
Table~\ref{tab:res_subgoal} shows the results.
It is seen that our method shows higher success rates in almost all of  the sub-goal categories. 

\subsection{Ablation Study}
We conduct an ablation test to validate the effectiveness of the components by incrementally adding each component to the proposed model. The results are shown in Table \ref{tab:ablation}. 

\begin{table}[h!]
    \begin{small}
    \begin{adjustwidth}{-0.05cm}{}
    \resizebox{0.49\textwidth}{!}{
        \begin{tabular}{@{}cccccc@{}}
        \toprule
        \multirow{3}{*}{Model} & \multicolumn{4}{c}{\bf Components} &  \multicolumn{1}{c}{\bf Validation} \\
        \cmidrule(r){2-5} \cmidrule(l){6-6}
        & {Instruction} & {Two-stage} & {Multi-view} & Mask  & \multicolumn{1}{c}{\multirow{2}{*}{Seen / Unseen}} \\
        & {Selection} & {Interpretation} & {Hier. Attn} & Decoder   \\

        \cmidrule(r){1-5} \cmidrule(l){6-6}
        
        1 & \xmark  & \xmark     &  \xmark    & \cmark & \multicolumn{1}{r}{2.8 / 0.5}  \\ 

        2 & \cmark  & \xmark     &       \xmark & \cmark & \multicolumn{1}{r}{12.9 / 2.9} \\  

        3 & \cmark  & \cmark     &  \xmark      & \cmark & \multicolumn{1}{r}{18.9 / 3.9} \\
        
        4 & \cmark  & \cmark     &  \xmark      & \xmark &  \multicolumn{1}{r}{3.8 / 0.7} \\

        5 & \cmark & \cmark     &       \cmark & \cmark &  \multicolumn{1}{r}{33.7 / 9.7} \\  
        \bottomrule
        \end{tabular}
    }
    \end{adjustwidth}

    \label{tab:dataset_comparison}
    \end{small}
    \vspace{-0.5em}
    \caption{
        \textbf{Ablation study for the components of the proposed model.}
        We report the success rate (Task score) on the validation seen and unseen splits. The {\xmark} mark denotes that a corresponding component is removed from the proposed model. 
    }
    \vspace{-1em}
    \label{tab:ablation}
\end{table}

\begin{figure*}[t!]
  \includegraphics[width=\linewidth]{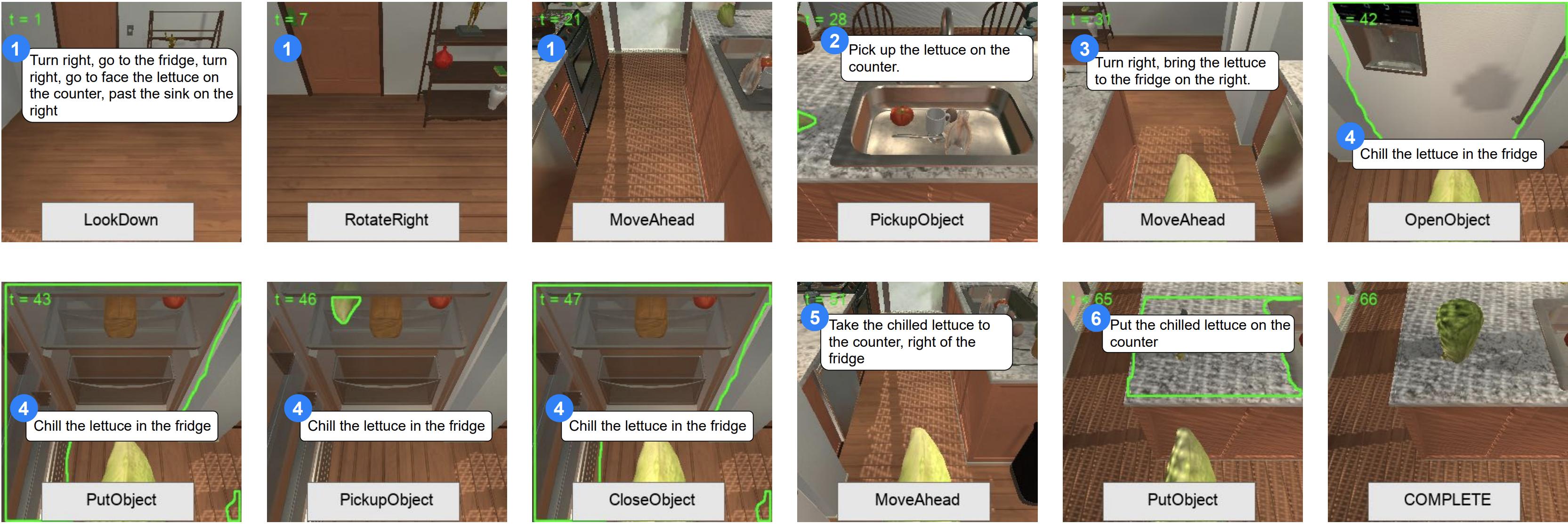}
\vspace{-0.2cm}
\caption{Our agent completes a \textbf{Cool} \& \textbf{Place} task ``\textit{Put chilled lettuce on the counter}" in an unseen environment.}
\label{fig:tasktype}
\vspace{-0.2cm}
\end{figure*}

The model variants 1-4 use a single-view input ($K=1$); they do not use multi-view inputs and the hierarchical attention method. Model 1 further discards the instruction decoder by replacing it with the soft-attention-based approach \cite{shridhar2020alfred}, which yields a different language feature $s_{\text{att}}$ at each timestep. Accordingly, $p^\text{io}_t$ and $p^\text{ia}_t$ are not fed to the mask/action decoders; we use $g^\text{m}_t = [s_\text{att}; h^\text{a}_t]$. These changes will make the method almost unworkable. Model 2 retains only the instruction selection module, yielding $s_{m_t}$. It performs much better than Model 1. Model 3 has the instruction decoder, which feeds $p^\text{io}_t$ and $p^\text{ia}_t$ to the subsequent decoders. It performs better than Model 2 by a large margin, showing the effectiveness of the two-stage method. 

Model 4 replaces the mask decoder with the counterpart of the baseline method \cite{shridhar2020alfred}, which upsamples a concatenated vector $[g^m_t; v_t]$ by deconvolution layers. This change results in inaccurate mask prediction, yielding a considerable performance drop. Model 5 is the full model. The difference from Model 3 is the use of multi-view inputs with the hierarchical attention mechanism. It contributes to a notable performance improvement, validating its effectiveness.

\subsection{Qualitative Results}

\subsubsection{Entire Task Completion}
Figure \ref{fig:tasktype} shows the visualization of how the agent completes one of the seven types of tasks. These are the results for the unseen environment of the validation set. Each panel shows the agent's center view with the predicted action and object mask (if existing) at different time-steps. {
See the appendix for more results.
}

\subsubsection{Mask Prediction for Sub-goal Completion}
\begin{figure}[h!]
\begin{subfigure}{.45\columnwidth}
  \centering
  \includegraphics[width=\linewidth]{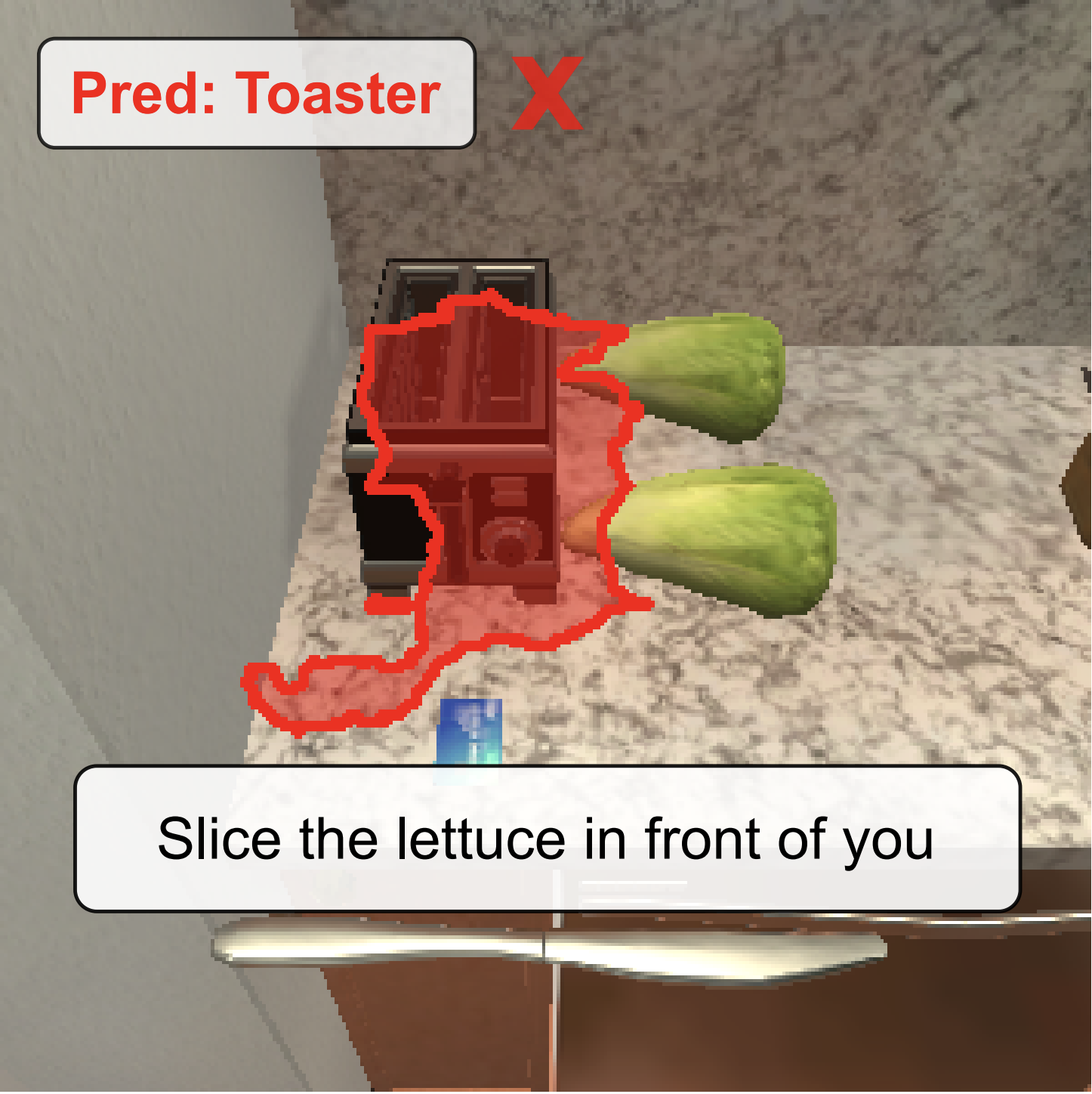}
  \caption{\cite{shridhar2020alfred}}
  \label{fig:sfig1}
\end{subfigure} \hfill 
\begin{subfigure}{.45\columnwidth}
  \centering
  \includegraphics[width=\linewidth]{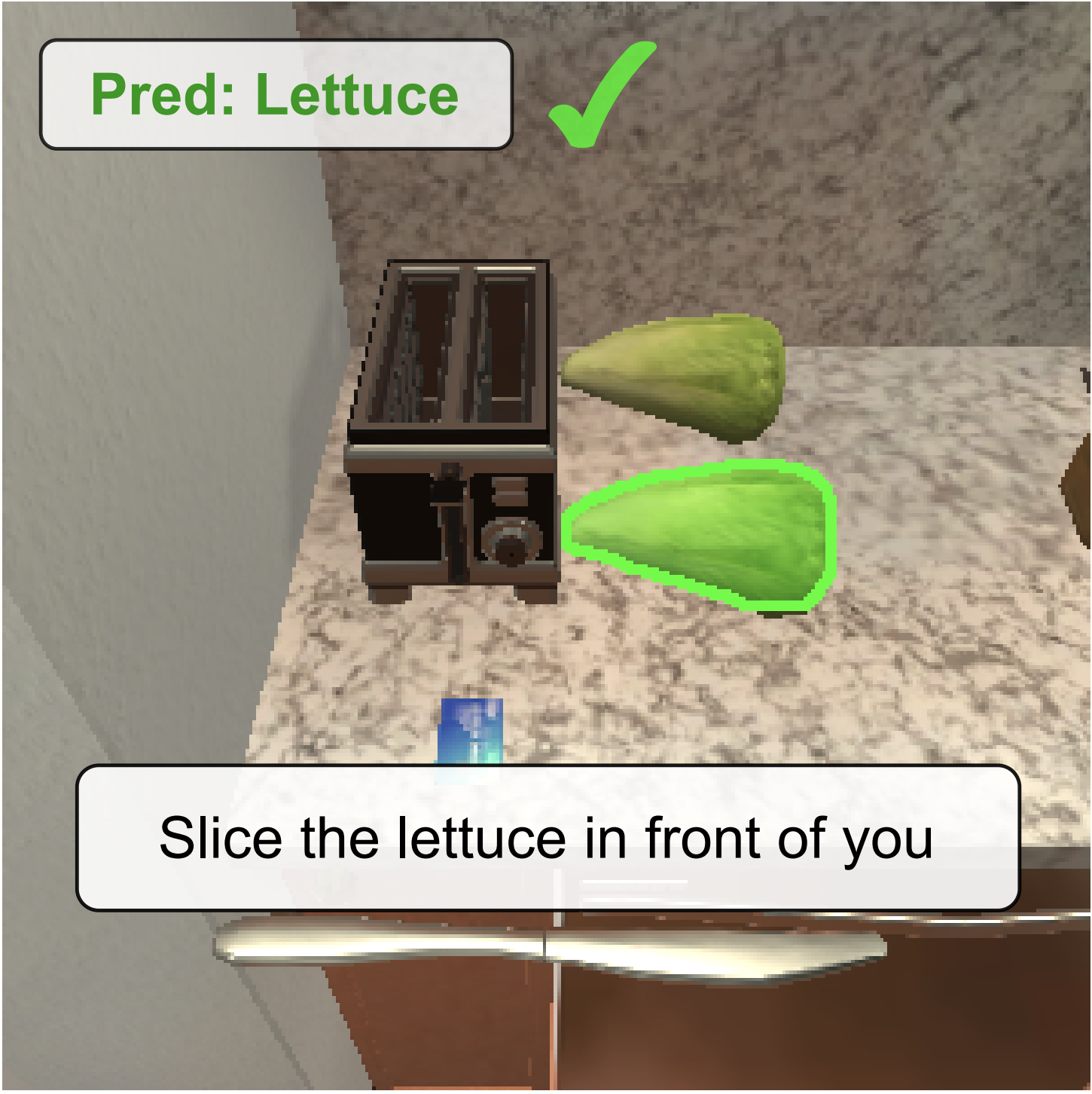}
  \caption{Ours}
  \label{fig:sfig2}
\end{subfigure}
\vspace{-0.1cm}
\caption{The prediction masks generated by  Shridhar \etal and our method where the agents are moved to the same location to accomplish {\bf Slice} sub-goal.}
\label{fig:vis_comparison}
\vspace{-0.2cm}
\end{figure}

Figure \ref{fig:vis_comparison} shows {
an example of the mask prediction} 
by the baseline \cite{shridhar2020alfred} and the proposed method.
It shows our method can predict a more accurate object mask when performing {\bf Slice} sub-goal. 
More examples are shown in the appendix.
Overall, our method shows better results, especially for difficult sub-goals like \textbf{Pickup}, \textbf{Put}, and \textbf{Clean}, for which a target object needs to be chosen from a wide range of candidates.

\section{Conclusion}

This paper has presented a new method for interactive instruction following tasks and applied it to ALFRED. The method is built upon several new ideas, including the explicit selection of one of the provided instructions, the two-stage approach to the interpretation of each instruction (i.e., the instruction decoder), the employment of the object-centric representation of visual inputs obtained by hierarchical attention from multiple surrounding views (i.e., the action decoder), and the precise specification of objects to interact with based on the object-centric representation (i.e., the mask decoder). The experimental results have shown that the proposed method achieves superior performances in both seen and unseen environments compared with all the existing methods. We believe this study provides a useful baseline framework for future studies.

\paragraph{Acknowledgments.} 
This work was partly supported by JSPS KAKENHI Grant Number 20H05952 and JP19H01110.

{\small
\bibliographystyle{named}
\bibliography{ijcai21}
}


\appendix

\section*{Appendix}
\setcounter{figure}{1}
\setcounter{table}{3}

\section{Additional Experimental Results}

\subsection{Performance by Task Type}

Table \ref{tab:tasktype} shows the success rates across the 7 task types achieved by the existing methods including ours on the validation set of ALFRED. It is seen that our method outperforms others by a large margin in both seen and unseen environments. 

\newcolumntype{d}{>{\columncolor{Gray}}c}
\begin{table}[h!]
    \centering
    \resizebox{0.48\textwidth}{!}{
        \begin{tabular}{@{}ldcdcdr@{}}
        \toprule
        \multirow{2}{*}{Task-Type}
                         & \multicolumn{2}{c}{\small\cite{shridhar2020alfred}}
                         & \multicolumn{2}{c}{\small\cite{singh2020moca}} 
                         & \multicolumn{2}{c}{\textbf{Ours}}
                         \\
                          \cmidrule(lr){2-3} \cmidrule(lr){4-5}  \cmidrule(lr){6-7}
                         & \multicolumn{1}{a}{Seen} & \multicolumn{1}{c}{Unseen} 
                         & \multicolumn{1}{a}{Seen} & \multicolumn{1}{c}{Unseen}
                         & \multicolumn{1}{a}{Seen} & \multicolumn{1}{c}{Unseen}
                         \\
                         
        \midrule
        {Pick \& Place}     & $7.0$ & $0.0$             & ${29.5}$ & $ {5.0}$
        & $\B{40.1}$ & $ \B{13.0}$\\

        {Stack \& Place}    & $0.9$ & $0.0$             & $5.2$ & $1.8$      
        & 
        $\B{17.4}$ & $ \B{11.9}$ \\
        
        {Pick Two}          & ${0.8}$ & ${0.0}$         & $11.2$ & $1.1$
        & 
        $\B{21.8}$ & $ \B{1.1}$ \\
        
        {Clean \& Place}    & $1.8$ & $0.0$             & 22.3 & 2.4        
        & 
        $\B{40.2}$ & $ \B{15.0}$ \\
        
        {Heat \& Place}     & $1.9$ & ${0.0}$           & ${15.8}$ & ${2.7}$  
        & 
        $\B{41.2}$ & $ \B{9.6}$ \\
        
        {Cool \& Place}     & ${4.0}$ & $0.0$           & $26.1$ & $0.7$         
        & 
        $\B{40.0}$ & $ \B{13.8}$ \\
        
        {Examine}           & ${9.6}$ & ${0.0}$       & $20.2$ & $\B{13.2}$
        & 
        $\B{34.4}$ & $ {12.9}$ \\
        \midrule
        {Average}    & $3.7$ & $0.0$  & ${18.6}$ & ${3.8}$          
        & 
        \B{33.6} & \B{11.0} \\
        \bottomrule
        \end{tabular}
    }

    \caption{
        \textbf{Success rate across 7 task types.} All values are in percentages. The agent is evaluated on the validation set. Highest values per split 
        are indicated in \B{bold}.
    }
    \label{tab:tasktype}
\end{table}

\subsection{Full Results of Ablation Tests}

Table \ref{tab:full_ablation1} shows the full results of the ablation test  reported in the main paper. We also provide additional results in Table \ref{tab:full_ablation2} with different activation functions (i.e. {sigmoid} or {softmax}) in the second step of the proposed hierarchical attention mechanism, and with different $K$'s; $K$ is selected from 1 (only `center' view), 3 (`center', `left', and `right' views), or 5 (`center', `left', `right', `up', and `down' views)). The results show that the use of \emph{gated-attention} in Eq.(5) (of the main paper) is essential. We also confirm the number of views also affect the success rate.

\begin{table*}[h]
    \centering
    \begin{small}
    \resizebox{0.8\textwidth}{!}{
        \begin{tabular}{@{}ccccrrrr@{}}
        \toprule
        \multicolumn{4}{c}{\bf Components} & \multicolumn{2}{c}{\bf Validation-Seen} & \multicolumn{2}{c}{\bf Validation-Unseen} \\
        \cmidrule(r){1-4} \cmidrule(lr){5-6} \cmidrule(l){7-8}
        {Instruction}  & {Two-stage} & {Multi-view} & {Mask} &
        \multicolumn{1}{c}{\multirow{2}{*}{Task}}   & \multicolumn{1}{c}{\multirow{2}{*}{Goal-Cond.}} &  \multicolumn{1}{c}{\multirow{2}{*}{Task}} &  \multicolumn{1}{c}{\multirow{2}{*}{Goal-Cond.}} \\
        {Selection} & {Interpretation} & {Hier. Attn} & {Decoder} & & & & \\

        \cmidrule(r){1-4} \cmidrule(lr){5-6} \cmidrule(l){7-8}
        \xmark & \xmark &     \xmark    & \cmark     & ${2.8}$ (${1.3}$)    & ${9.7}$  (${6.5}$)     & ${0.5}$ (${0.2}$)    & ${9.2}$ (${5.4}$)  \\ 
        
        \cmark &    \xmark  & \xmark     &  \cmark    & ${12.9}$ (${9.4}$)    & ${21.6}$  (${17.3}$)     & ${2.9}$ (${1.6}$)    & ${13.1}$ (${9.4}$) \\

        \cmark &    \cmark  & \xmark     &  \cmark      &${18.9}$ (${13.9}$)    & ${26.8}$  (${21.9}$)     & ${3.9}$ (${2.5}$)    & ${15.3}$ (${10.9}$) \\
        
        \cmark &    \cmark  & \cmark     &  \xmark      &${3.8}$ (${2.4}$)& ${14.9}$ (${11.2}$) & ${0.7}$ (${0.3}$)& ${10.4}$ (${6.9}$) \\ 
        
        \cmark & \cmark & \cmark     &       \cmark         & {${33.7}$ (${28.4}$)}    & ${43.1}$  (${38.0}$)    & ${9.7}$ (${7.3}$)    & ${23.1}$ (${18.1}$) \\ 
        \bottomrule
        \end{tabular}
    }
    \end{small}
    \caption{
        Results of an ablation test for examining the effectiveness of each component of the proposed model. The path weighted scores   are reported in the parentheses.
    }
    \label{tab:full_ablation1}
\end{table*}

\begin{table*}[h]
    \centering
    \begin{small}
    \resizebox{0.8\textwidth}{!}{
        \begin{tabular}{@{}ccccrrrr@{}}
        \toprule
        \multicolumn{4}{c}{\multirow{2}{*}{\bf Configurations}} & \multicolumn{2}{c}{\bf Validation-Seen} & \multicolumn{2}{c}{\bf Validation-Unseen} \\
        \cmidrule(lr){5-6} \cmidrule(l){7-8}
        & & & & \multicolumn{1}{c}{{Task}}   & \multicolumn{1}{c}{{Goal-Cond.}} &  \multicolumn{1}{c}{{Task}} &  \multicolumn{1}{c}{{Goal-Cond.}} \\

        \cmidrule(r){1-4} \cmidrule(lr){5-6} \cmidrule(l){7-8}
        \multicolumn{2}{l}{\multirow{2}{*}{Activation Function}} &   \multicolumn{2}{c}{\hspace{0.5cm}softmax}     &  ${11.9}$ (${9.3}$)    & ${20.8}$ (${17.3}$)  & ${4.1}$ (${2.2}$)    & ${14.0}$  (${10.2}$) \\ 
        
        & &  \multicolumn{2}{c}{\hspace{0.5cm}\textbf{sigmoid}}    & {${33.7}$ (${28.4}$)}    & ${43.1}$  (${38.0}$)    & ${9.7}$ (${7.3}$)    & ${23.1}$ (${18.1}$) \\ 
        
        \cmidrule(r){1-4} \cmidrule(lr){5-6} \cmidrule(l){7-8}
        & &  \multicolumn{2}{c}{\hspace{0.5cm}center}      &${18.9}$ (${13.9}$)    & ${26.8}$  (${21.9}$)     & ${3.9}$ (${2.5}$)    & ${15.3}$ (${10.9}$) \\
        
        \multicolumn{2}{l}{{Ego-centric views}}  & \multicolumn{2}{c}{\hspace{0.5cm}center, left, right}      &${25.9}$ (${21.2}$)& ${34.4}$ (${30.0}$) & ${6.2}$ (${3.8}$)& ${17.0}$ (${12.3}$) \\ 
        
        & & \multicolumn{2}{c}{\hspace{0.5cm}\textbf{center, left, right, up, down} \hspace{3cm}}       & {${33.7}$ (${28.4}$)}    & ${43.1}$  (${38.0}$)    & ${9.7}$ (${7.3}$)    & ${23.1}$ (${18.1}$) \\ 
        \bottomrule
        \end{tabular}
    }
    \end{small}
    \caption{
        Results of experiments comparing activation functions in the module for aggregating and encoding multi-view visual inputs. 
        The path weighted scores are reported in the parentheses.
    }
    \label{tab:full_ablation2}
    \vspace{-0.5cm}
\end{table*}

\section{Qualitative Results}

\subsection{Mask Prediction for Sub-goal Completion}

\begin{figure*}[t!]
    \centering
    \includegraphics[width=0.85\linewidth]{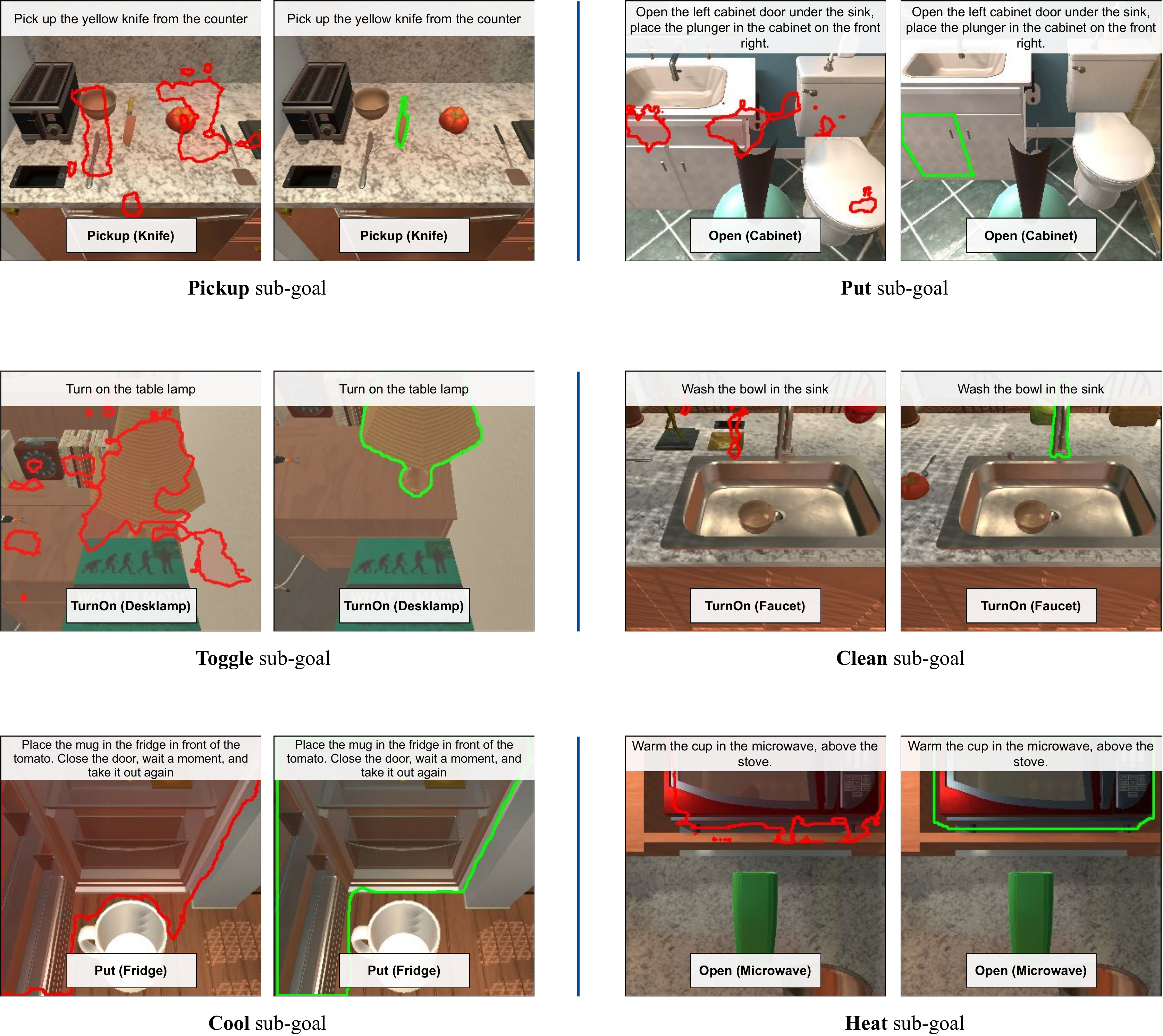}
    \caption{Examples of the object masks predicted by the baseline method [Shridhar \etal, 2020] (left on each panel) and our method (right) when the corresponding agents are at the same location and perform one of the six manipulation sub-goals.}
    \label{fig:all_comparison}
\end{figure*}

Figure \ref{fig:all_comparison} shows examples of mask predictions by the baseline \cite{shridhar2020alfred} and the proposed method for different sub-goals. 
Overall, it is seen that our method can predict more accurate object masks. It shows better results especially for difficult sub-goals like \textbf{Pickup}, \textbf{Put}, and \textbf{Clean}, where a target object needs to be chosen from a wide range of candidates.

\subsection{Entire Task Completion}

\begin{figure*}[h]
\centering
\includegraphics[width=\linewidth]{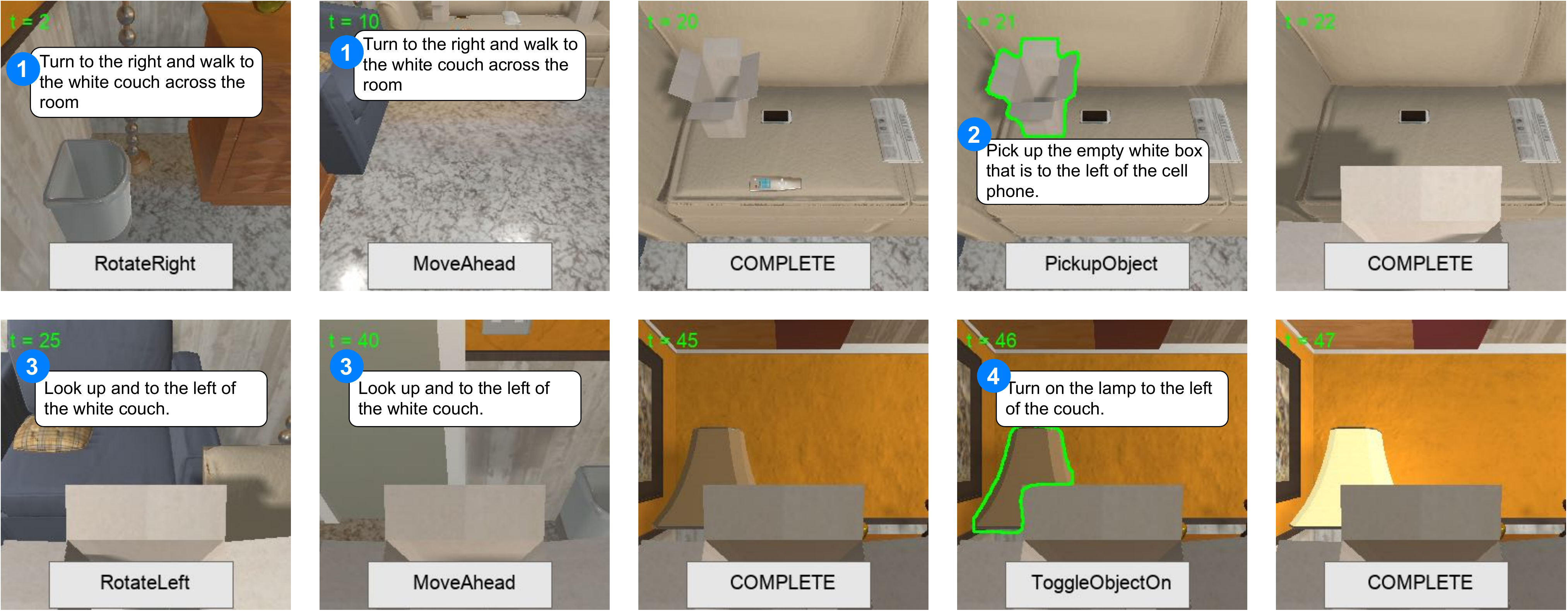}
\vspace{-0.1cm}
\caption{Our agent completes an \textbf{Examine} task ``\textit{Examine an empty box by the light of a floor lamp}" in an unseen environment.}
\label{fig:tasktype0}
\vspace{-0.1cm}
\end{figure*}

\begin{figure*}[t!]
  \includegraphics[width=\linewidth]{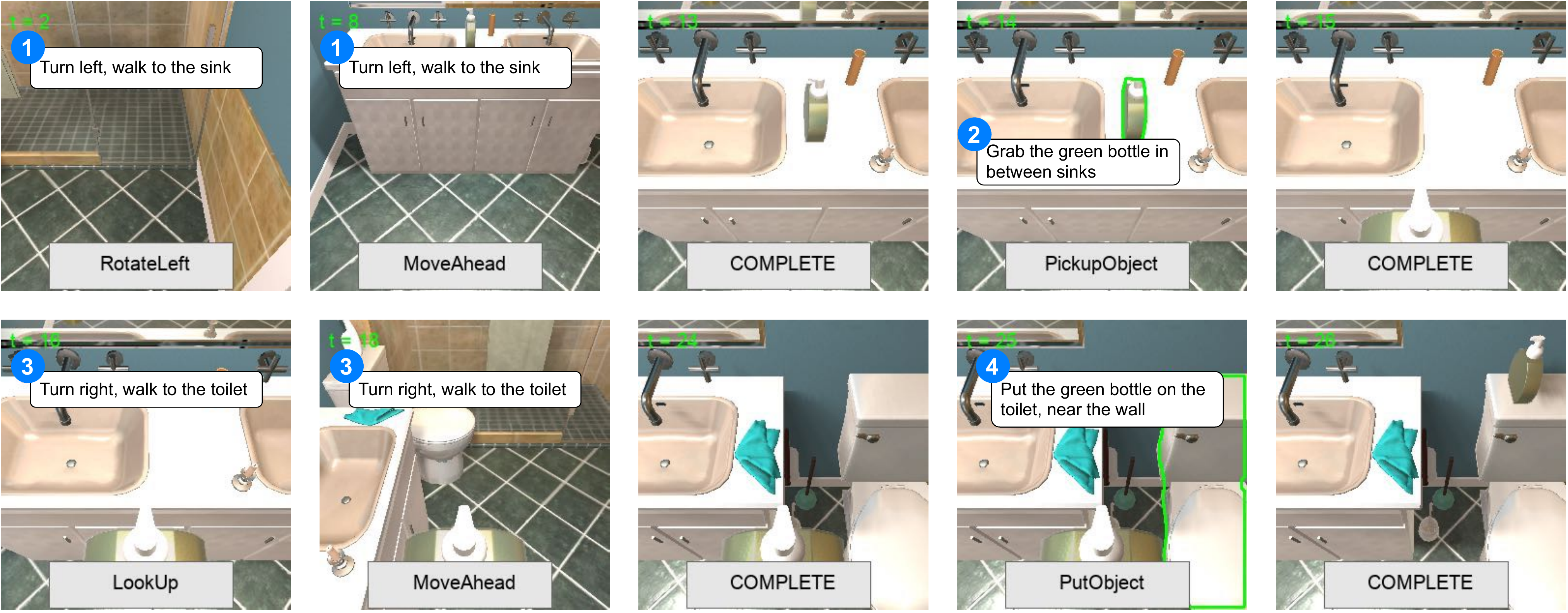}
\vspace{-0.1cm}
\caption{Our agent completes a \textbf{Pick} \& \textbf{Place} task ``\textit{Place the green bottle on the toilet basin}" in an unseen environment.}
\label{fig:tasktype1}
\vspace{-0.1cm}
\end{figure*}

\begin{figure*}[t!]
  \includegraphics[width=\linewidth]{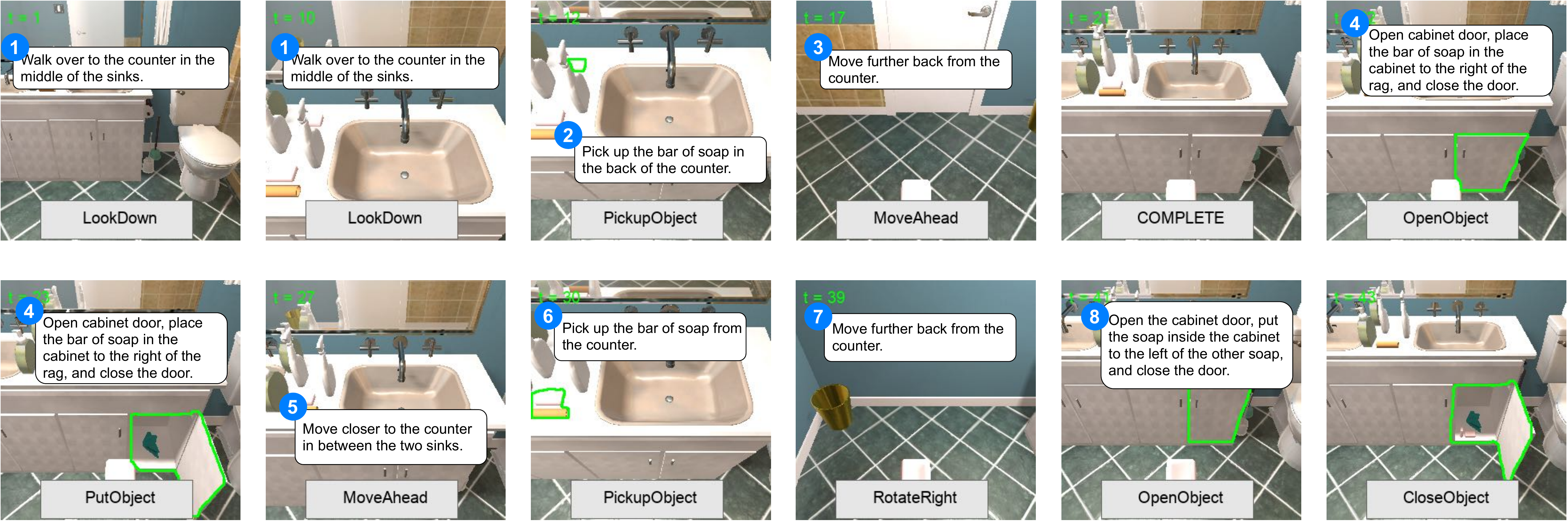}
\vspace{-0.1cm}
\caption{Our agent completes a \textbf{Pick Two} \& \textbf{Place} task ``\textit{To move two bars of soap to the cabinet}" in an unseen environment.}
\label{fig:tasktype2}
\vspace{-0.1cm}
\end{figure*}

\begin{figure*}[t!]
  \includegraphics[width=\linewidth]{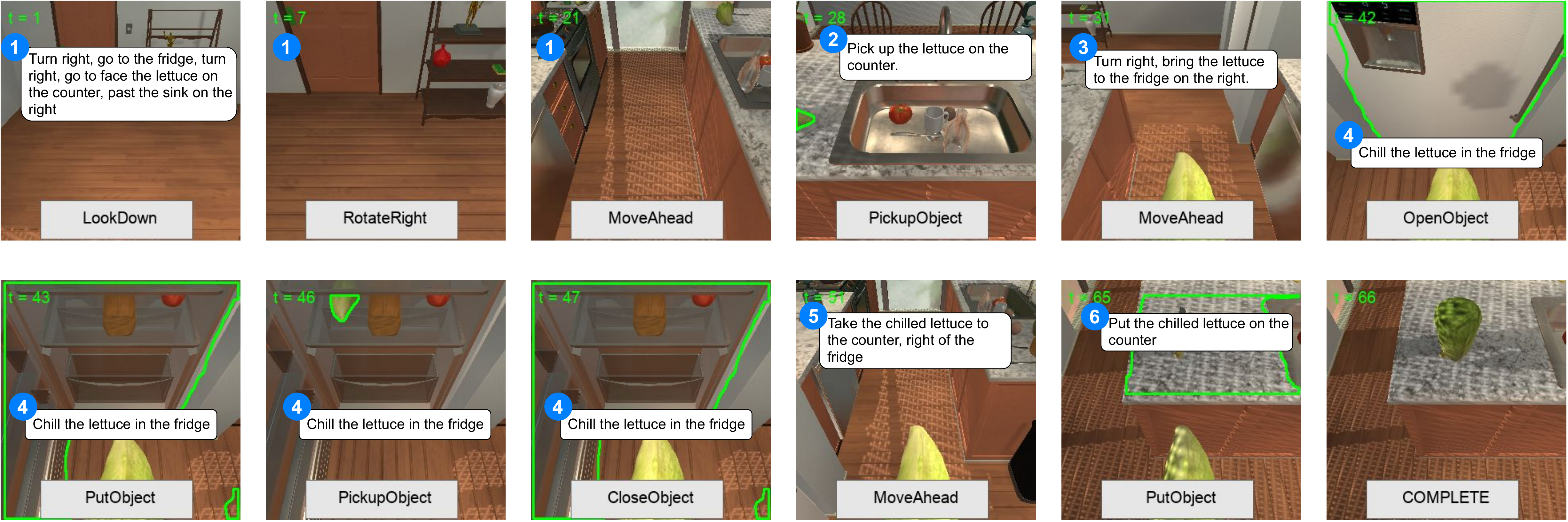}
\vspace{-0.1cm}
\caption{Our agent completes a \textbf{Cool} \& \textbf{Place} task ``\textit{Put chilled lettuce on the counter}" in an unseen environment.}
\label{fig:tasktype3}
\vspace{-0.1cm}
\end{figure*}

\begin{figure*}[t!]
  \includegraphics[width=\linewidth]{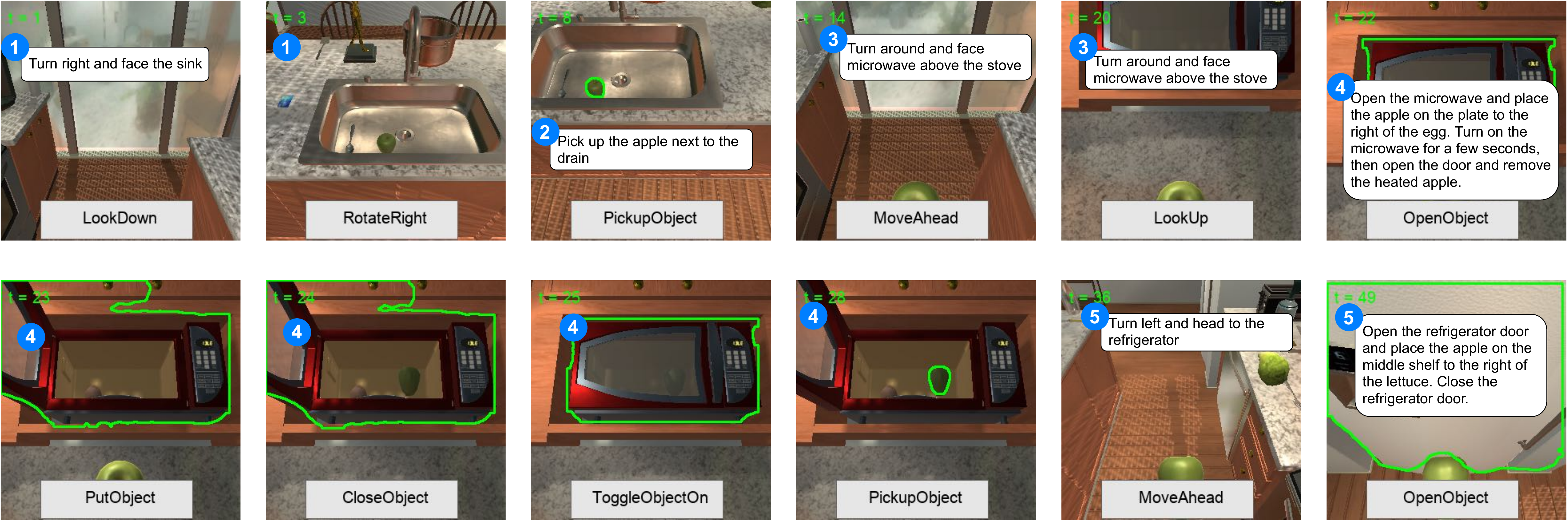}
\vspace{-0.1cm}
\caption{Our agent completes a \textbf{Heat} \& \textbf{Place} task ``\textit{Put a heated apple next to the lettuce on the middle shelf in the refrigerator}" in an unseen environment.}
\label{fig:tasktype4}
\vspace{-0.1cm}
\end{figure*}

\begin{figure*}[t!]
  \includegraphics[width=\linewidth]{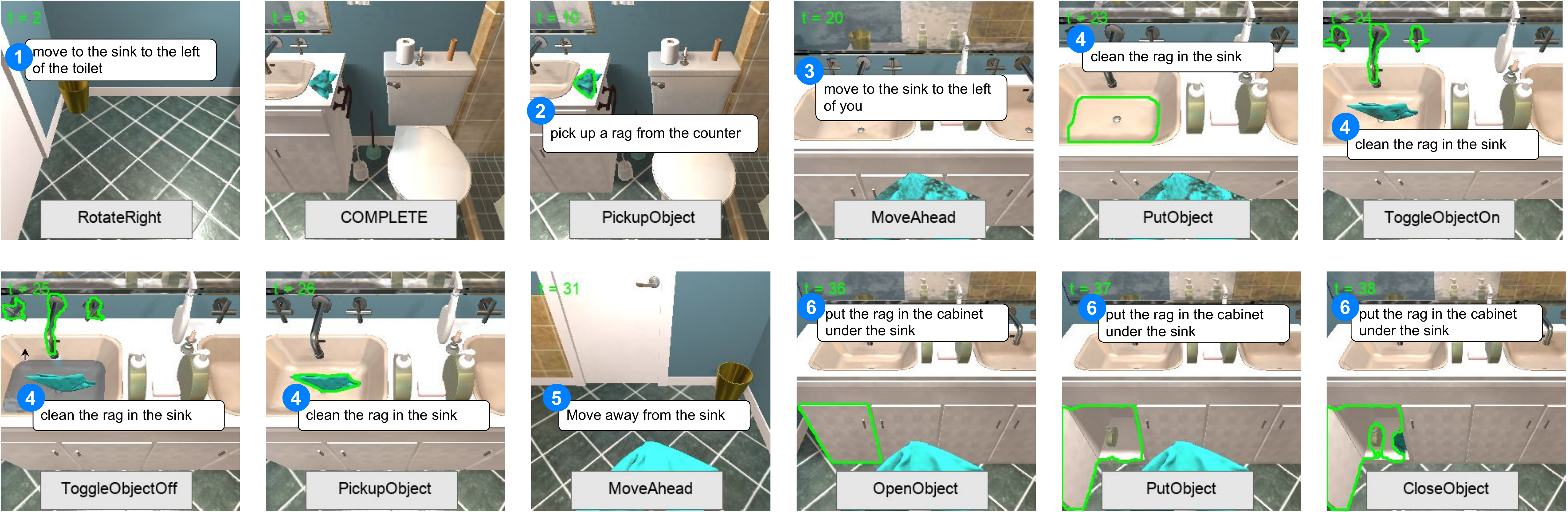}
\vspace{-0.1cm}
\caption{Our agent completes a \textbf{Clean} \& \textbf{Place} task ``\textit{Put a cleaned rag in the cabinet under the sink}" in an unseen environment.}
\label{fig:tasktype5}
\vspace{-0.1cm}
\end{figure*}

\begin{figure*}[t!]
  \includegraphics[width=\linewidth]{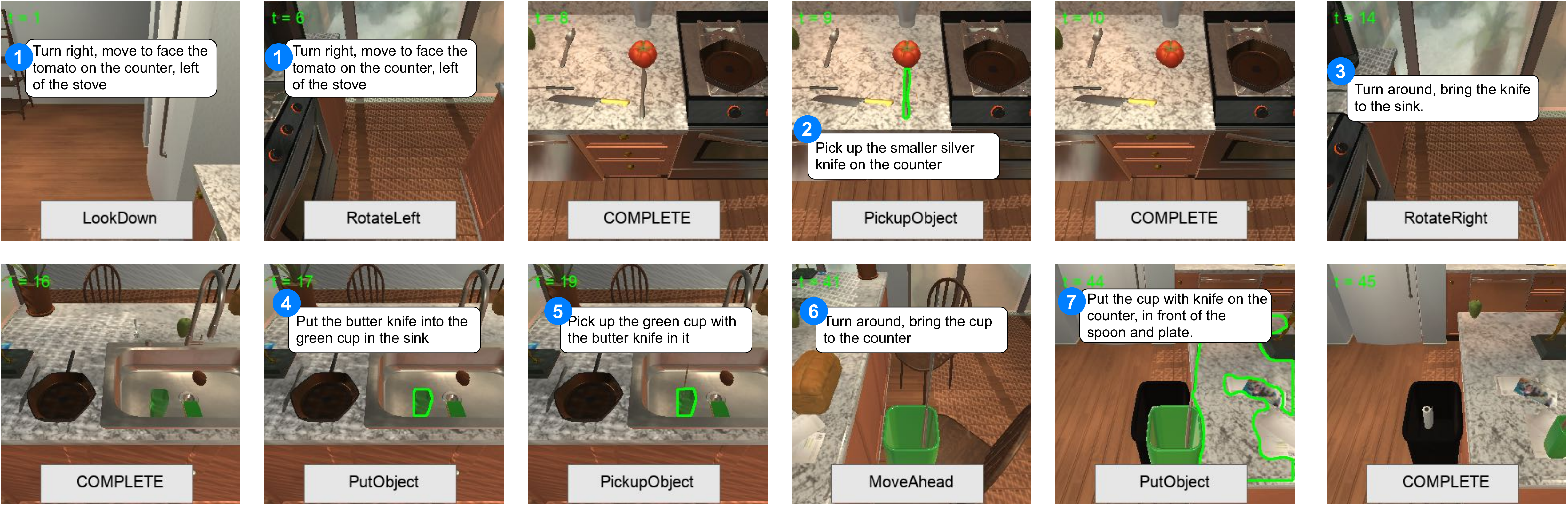}
\vspace{-0.1cm}
\caption{Our agent completes a \textbf{Stack} \& \textbf{Place} task ``\textit{Put a cup with knife in it, on the counter}" in an unseen environment.}
\label{fig:tasktype6}
\vspace{-0.1cm}
\end{figure*}

Figures \ref{fig:tasktype0}-\ref{fig:tasktype6} show  example visualization of how the agent completes the seven types of tasks. These are the results for the unseen environments of the validation set. Each panel shows the agent's center view with the predicted action and object mask (if existing) at different timesteps. 

We also provide seven video clips as independent files, which contain several examples of the agent's entire task completion for seven above task instances in unseen environments. 


\section{Analyses of Failure Cases}

We analyze the failure cases of our method using the results on the validation splits.
We categorize them into navigation failures and manipulation failures. 

\subsection{Navigation Failures} 

It is seen from the sub-goal results of Table 2 in the main paper that the \textbf{Goto} sub-goal is the most challenging. Failures with it tend to make it hard to complete the entire goal, since they will inevitably affect the subsequent actions to take. We think there are three major cases for the navigation failures.  

The first case, which occurs most frequently, is that 
the agent follows a navigation instruction and reaches a position that should be fine as far as the instruction goes; nevertheless, it is not the right position for the next manipulation action to take. For instance, following the instruction ``Go to the table,'' the agent goes to the table. The next instruction is `` Pickup the remote control at the table,'' but the remote lies on the other side of the table. This is counted as a failure of completing the \text{Goto} sub-goal. 

The second case is when the instructions are either  abstract or misleading. An example is that when the agent has to take several left and right turns together with multiple \texttt{MoveAhead} steps to reach the destination, e.g., a drawer, the provided instruction is simply``Go to the drawer.''

The third case, which occurs less frequently, is that
while there is an obstacle in front of the agent, e.g., wall, it attempts to take the {\tt MoveAhead} action. This occurs because of the lack of proper visual inputs. 
This is demonstrated by the fact that when we reduce the number of views, the task success rate drops significantly, as shown in the second block of Table \ref{tab:full_ablation2}.

\subsection{Manipulation Failures} 

As shown in Table 2, after it has moved to the ideal position right before performing any interaction sub-goals (i.e., all the sub-goals but \textbf{Goto}), the agent can manipulates objects with high success rates of 91\% and 69\% in the seen and unseen environments, respectively. 

However, the success rates for completing the Goto sub-goal in the seen and unseen environments are only 59\% and 39\%, respectively. Therefore, the primary cause of the manipulation failures is that the agent cannot find the target object becuase it fails to reach the right destination due to a navigation failure. 


Even if the agent has successfully navigated to the right destination, it can fail to detect the target object. This seem to happen mostly because the object is either too small or indistinguishable from the surroundings. The agent tends to fail to detect, for example, a small knife placed on the steel/metal-made sink of the same color. 

The agent also fails to detect an object that has not seen in the training. This is confirmed by the fact that the performance drops considerably in unseen environments for some interaction sub-goals (including \textbf{Put}, \textbf{Clean}, and \textbf{Toggle}). 
There are also a small number of cases where failures are attributable to bad instructions, e.g., incorrect statement of objects.

\section{Further Details of Implementation}

Table \ref{tab:hyperparams} summarizes the hyperparameters used in our experiments. We perform all the experiments on a GPU server that has four Tesla V100-SXM2 of 16GB memory. It has Intel(R) Xeon(R) Gold 6148 CPU @ 2.40GHz of 20 cores with the RAM of 240GB memory. We use Pytorch version 1.6 \cite{NEURIPS2019_9015}. As for the Mask R-CNN model, we train it in advance, separately from the main model, for 15 epochs with the learning rate $10e-3$ and halved at the epoch 5, 8, and 12. We train models with batch size of 32 and 4 workers per GPU.

\begin{table}[h]
    \centering
    \caption{Hyperparamters used in the training.}
    \begin{tabular}{ll}
    \toprule
    Hyperparameter          &  Value \\
    \midrule
    $\beta_1$ in Adam & 0.9 \\
    $\beta_2$ in Adam & 0.999 \\
    $\epsilon$ in Adam & $1\mathrm{e}{-8}$  \\
    Number of workers per GPU & 4 \\
    Batch size & 32 \\
    \bottomrule
    \end{tabular}
    \label{tab:hyperparams}
\end{table}

\section{Entry Submission to the ALFRED Embodied AI Challenge 2021}

\begin{figure}[h]
\centering
  \includegraphics[width=0.85\linewidth]{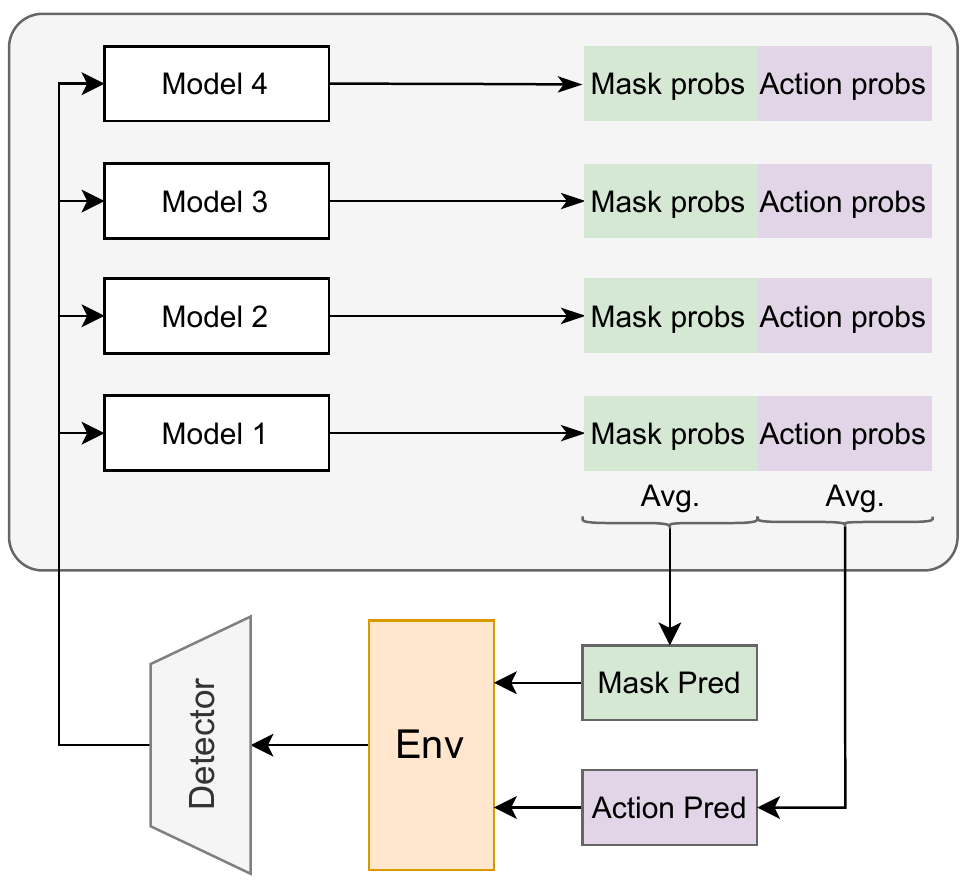}
\caption{Illustration of ensemble during the ALFRED evaluation.}
\label{fig:ensemble}
\end{figure}

In this section, we describe the details of our entry submission to the ALFRED Embodied AI Challenge 2021\footnote{https://askforalfred.com/EAI21}, which is organized in conjuction with the CVPR 2021.

To further improve the performance of our model, we create an ensemble of 4 models with some differences, from initializations with different random seeds to whether to use self-attention mechanism in the mask decoder. We train these 4 models independently with the same pretrained Mask R-CNN, whose weights fixed during the training phase. Refer to these above sections for training and implementation details.

During the evaluation, a shared detector Mask R-CNN is to extract the features from visual inputs, which are forwarded into all the models. Each model will ouput the probability distributions over the actions and objects of interest if any. We then take average these probabilities and select the actions and objects of highest probabilities. Figure \ref{fig:ensemble} illustrates the use of ensemble during the inference.

\begin{table}[!h]
    \centering
    \resizebox{0.48\textwidth}{!}{
        \begin{tabular}{@{}laarr@{}}
            \toprule
            \multicolumn{1}{l}{\multirow{3}{*}{Model}} & \mcc{4}{\textbf{Test}} \\
            {} & \mcc{2}{\textit{Seen}} & \mcc{2}{\textit{Unseen}}  \\
            {} & \multicolumn{1}{b}{Task} & \multicolumn{1}{b}{Goal-Cond} & \multicolumn{1}{c}{Task} & \multicolumn{1}{c}{Goal-Cond} \\

            \cmidrule{1-5}
            \multicolumn{5}{l}{\bf 5 views} \\
            \multicolumn{1}{l}{Our single model} & ${29.16}$ (${24.67}$) & ${38.82}$ (${34.85}$)   & ${8.37}$ (${5.06}$) & ${19.13}$ (${14.81}$) \\[1pt]
            \multicolumn{1}{l}{Our ensemble $\diamond$} & $\B{30.92}$ ($\B{25.90}$) & $\B{40.53}$ ($\B{36.76}$)   & $\B{9.42}$ ($\B{5.60}$) & $\B{20.91}$ ($\B{16.34}$) \\[1pt]
            \bottomrule
        \end{tabular}
    }
    \caption{\textbf{Task and Goal-Condition Success Rate.}
        For each metric, the corresponding path weighted metrics are given in (parentheses). Our entry submission to the EAI Challenge 2021 is denoted with $\diamond$.
    }
    \label{tab:ensemble_results}
\end{table}

Table \ref{tab:ensemble_results} shows the performance of our single model and its corresponding ensemble. It is seen that ensembling increases the success rate of the agent on both seen and unseen environments in comparison with a single model. 

It is noted that the evaluation speed is slowed down proportionally when capturing multiple ego-centric views from the agent's camera. It is due to the fact that the current version of AI2Thor does not support direct acquisition of multiple/panoramic views; we thus have to move agents into different view points to capture such corresponding views. Our best-performing results from multi-view models suggest that it be of importance to improve the benchmark which is capable of obtaining multiple/panoramic views directly. Also noted that using ensemble during evaluation requires running all of its models simultaneously. Therefore, it leads to a proportional increase in memory and compute. 

\end{document}